%% file: main.tex
\documentclass{article}

\input{macros}

\makeatletter
\newcommand{\abstracttext}[1]{\def\@abstract{#1}}

\renewcommand{\maketitle}{%
    \begingroup
    \renewcommand{\thefootnote}{\fnsymbol{footnote}}  

  \begin{center}
        {\LARGE \bfseries \@title \par}
        \vspace{2em}

        \begin{tabular*}{\textwidth}{@{\extracolsep{\fill}} c c c c}
            \shortstack{\textbf{Atul Ganju}\\Cornell University\\ag2222@cornell.edu} &
            \shortstack{\textbf{Shashaank Aiyer\footnotemark[1]}\\Cornell University\\saa244@cornell.edu} &
            \shortstack{\textbf{Ved Sriraman\footnotemark[1]}\\Cornell University\\vs346@cornell.edu} &
            \shortstack{\textbf{Karthik Sridharan}\\Cornell University\\sridharan@cs.cornell.edu}
        \end{tabular*}

        \vspace{1em}
        {\@date \par}
    \end{center}

    \footnotetext[1]{Equal contribution.}
    \endgroup  

    \vspace{1em}
    \begin{center}
        \begin{minipage}{0.85\textwidth}
            \small
            \setlength{\parindent}{1.5em}
            \justifying
            \@abstract
        \end{minipage}
    \end{center}
    \vspace{2em}
}
\makeatother

\title{Active Learning via Regression Beyond Realizability}
\date{May 24, 2025}
\abstracttext{\input{section_0/abstract}
}

\begin{document}
\maketitle
\setcounter{footnote}{0}
\renewcommand{\thefootnote}{\arabic{footnote}}

\input{section_1/introduction}
\input{section_2/preliminaries}
\input{section_3/prelude}
\input{section_4/active_learning}
\input{section_5/discussion}

\newpage

\bibliography{references}
\newpage

\appendix
\input{section_6/appendix}

\bibliographystyle{alpha}

\end{document}

%% file: macros.tex
\usepackage{titling,kantlipsum}
\usepackage{amsfonts,amsthm,amsmath,amssymb}
\usepackage[margin=1in]{geometry}
\usepackage{xcolor}
\usepackage{graphicx}
\usepackage{subcaption}
\usepackage{enumitem}
\usepackage{mdframed}
\usepackage{authblk}
\usepackage{parskip}
\usepackage{bbm}
\usepackage{ragged2e}
\usepackage{algorithm}
\usepackage{algpseudocode}
\usepackage{setspace}
\usepackage[colorlinks, linkcolor=black,citecolor=black,urlcolor = black, bookmarks=true]{hyperref}
\usepackage[nameinlink, capitalize]{cleveref}
\usepackage{booktabs} 
\hypersetup{
	colorlinks,
	linkcolor={red!75!black},
	citecolor={blue!75!black},
	urlcolor={red!75!black},
}
\allowdisplaybreaks

\newtheorem{theorem}{Theorem}[section]
\newtheorem{claim}{Claim}
\newtheorem{assumption}{Assumption}
\newtheorem{lemma}{Lemma}
\newtheorem{definition}[theorem]{Definition}
\newtheorem{corollary}{Corollary}

\newtheorem{example}{Example}

\newtheorem{proposition}{Proposition}

\DeclareMathOperator*{\argmin}{argmin}
\DeclareMathOperator*{\argmax}{argmax}

\newcommand*\calF{\mathcal{F}} 

\newcommand*\calH{\mathcal{H}}
\newcommand*\calX{\mathcal{X}}
\newcommand*\calY{\mathcal{Y}}
\newcommand*\calD{\mathcal{D}}
\newcommand*\calO{\mathcal{O}}
\newcommand*\calN{\mathcal{N}}
\newcommand*\bbR{\mathbb{R}}
\newcommand*\bbE{\mathbb{E}}
\newcommand*\bbP{\mathbb{P}}
\newcommand*\bfe{\mathbf{e}}
\newcommand*\ind{\mathbbm{1}}

\DeclareMathOperator*{\Uniform}{Unif}

\newcommand*\comp{\mathsf{comp}}
\newcommand*\margin{\mathsf{margin}}
\newcommand*\gap{\mathsf{gap}}
\newcommand*\class{c}
\newcommand*\Oracle{\mathsf{Alg}^{\mathsf{OR}}_{\ell_\Phi}}
\newcommand*\Pdim{\text{Pdim}(\calF)}
\newcommand*\Eclass{\mathcal{E}_{0\textrm{-}1}}

\newcommand*\Esq{\mathcal{E}_{sq}}

%% file: section_0/abstract.tex
We present a new active learning framework for multiclass classification based on surrogate risk minimization that operates beyond the standard realizability assumption. Existing surrogate-based active learning algorithms crucially rely on realizability—the assumption that the optimal surrogate predictor lies within the model class—limiting their applicability in practical, misspecified settings. In this work we show that under conditions significantly weaker than realizability, as long as the class of models considered is convex, one can still obtain a label and sample complexity comparable to prior work. Despite achieving similar rates, the algorithmic approaches from prior works can be shown to fail in non-realizable settings where our assumption is satisfied. Our epoch-based active learning algorithm departs from prior methods by fitting a model from the full class to the queried data in each epoch and returning an improper classifier obtained by aggregating these models.

%% file: section_1/introduction.tex
\section{Introduction}\label{sec:intro}

We study active learning for multi-class classification in the statistical learning framework, where the learner has access to a large pool of unlabeled data and can query labels from an oracle. The objective is to learn an accurate classifier while minimizing the number of label queries—that is, to simultaneously achieve low excess classification risk and label complexity.

A promising recent approach to active learning is to reduce the problem to regression \cite{krishnamurthy, sekhari2023selective, zhu2022efficientactivelearningabstention, hanneke_2019}. Rather than directly optimizing the classification loss, these methods relax the problem to one of minimizing a convex, differentiable surrogate loss over a class of real-valued functions, thereby shifting the objective to approximating a real-valued target function whose associated classifier is expected to perform well under the original classification loss. This formulation enables the hypothesis class to encode additional structure—such as smoothness, margin, or regularization—that reflects assumptions about the target function, ultimately enabling the learning of richer, more expressive model classes. Furthermore, the differentiability of the surrogate facilitates the use of efficient gradient-based optimization techniques, which are essential for training complex models in modern machine learning systems.

While surrogate-based methods have become a cornerstone of modern learning systems due to their flexibility and compatibility with gradient-based optimization, their theoretical guarantees—both in the active setting \cite{krishnamurthy, sekhari2023selective, zhu2022efficientactivelearningabstention, hanneke_2019} and even in the passive setting \cite{bartlett2006convexity}—have critically relied on the \emph{realizability} assumption: that the minimizer of the surrogate risk lies within the function class accessible to the learner. However, assuming that a setting is realizable is often too strong to hold in
practice. To address this limitation, we develop a regression-based active learning algorithm that matches the accuracy and label complexity of prior realizability-based methods, while operating under conditions strictly milder than realizability.

To this end, we first establish a passive learning result that extends the realizability-based theory of \cite{bartlett2006convexity}. Specifically, we show that for convex function classes, there exists a broad family of classification-calibrated surrogate losses for which the excess classification risk can be upper bounded by the excess surrogate risk, provided a structural condition holds: namely, that the bias of the best-in-class function is lower bounded by a non-decreasing function of the bias of the surrogate risk minimizer. To demonstrate the generality of this condition, we construct an example in which realizability, and even approximate realizability, fails, yet our condition remains satisfied. Under additional distributional assumptions—such as those introduced by Massart and Tsybakov—our analysis yields sharper convergence rates, analogous to those obtained in the realizable setting by \cite{bartlett2006convexity}.

While this extension is conceptually simple, it serves as the foundation for our main contribution. Specifically, we design a novel, improper active learning algorithm for multi-class classification. Our algorithm achieves guarantees comparable to existing realizability-dependent, surrogate-based methods yet operates under a natural extension of our passive learning assumption that is strictly weaker than realizability. In contrast, existing algorithms such as those of \cite{hanneke_2019, zhu2022efficientactivelearningabstention} fundamentally rely on realizability, and their analyses break down in the misspecified setting. Our algorithm is the first to offer surrogate-based active learning guarantees in such generality.

\textbf{Related Work:}
The study of active learning initially focused on the noise-free setting, where the label distribution is assumed to be perfectly consistent with some classifier in a known hypothesis class \cite{freund_1997, dasgupta_2004, dasgupta_2005, hanneke_2015}. Foundational contributions in this regime include the general label complexity bounds for arbitrary hypothesis classes established by \cite{dasgupta_2005}, and the minimax rates derived by \cite{hanneke_2015}.

As the noise-free setting was increasingly recognized as too restrictive, attention shifted to the agnostic setting, where no assumptions are made on the label distribution. The work of \cite{balcan2006agnostic} introduced the disagreement-based framework and proposed the first algorithm for agnostic active learning with a nontrivial label complexity bound, as was later analyzed in \cite{hanneke2007bound}. However, subsequent work showed that in the fully agnostic regime, active learning cannot, in general, significantly outperform passive learning \cite{kaariainen2006active, hsu2010algorithms, hanneke2014theory}.

To circumvent this limitation, a line of research focused on characterizing label complexity as a function of distributional noise. This led to significant improvements in label complexity under structured noise models \cite{castro2008minimax, hanneke2009adaptive, koltchinskii2009rademacher, hanneke2011rates, locatelli2017adaptivitynoiseparametersnonparametric}. A particularly notable result by \cite{hanneke_2015} established minimax label complexity bounds for a broad class of noise models, demonstrating that under Tsybakov noise, active learning can provably outperform passive learning for any VC class.

Although these results establish strong theoretical foundations, they rely on idealized algorithmic primitives such as classification oracles and are limited to relatively simple hypothesis classes. To address these shortcomings, a series of works explored surrogate loss minimization as an alternative framework for active learning, with several promising results under the realizability assumption \cite{audibert_2007, minsker2012plugin, hanneke_2019, krishnamurthy, zhu2022efficientactivelearningabstention}.

Among these, the work of \cite{hanneke_2019} is most directly related to ours. Their key insight is that once the sign of the optimal predictor is known at a given point, its exact value becomes irrelevant for classification. This allows their algorithm to focus exclusively on the region where the sign remains uncertain. In particular, it operates by maintaining a version space that is progressively refined by optimizing the empirical surrogate risk over this region of uncertainty. As such, their algorithm obtains a good approximation of the optimal predictor and therefore is able to infer the sign at additional inputs with increasing confidence. Although this insight still holds in the misspecified setting, their algorithmic approach fundamentally relies on the realizability assumption. In fact, the authors remark that identifying a concise, \textit{a priori} condition under which their guarantees would extend to the non-realizable case would require a substantially different algorithmic approach. In this respect, our algorithm departs significantly from theirs: rather than refining a single version space, we employ an improper learning strategy that constructs a piecewise aggregation of classifiers, each operating on a region of the input space where it maintains high confidence.

Also relevant are the works of \cite{bartlett2006convexity}, \cite{liang2015learning}, and \cite{zhu2022efficientactivelearningabstention}, each of which contributes essential mathematical ingredients to our analysis. The results of \cite{bartlett2006convexity} provide the foundation for bounding classification excess risk in terms of surrogate excess risk. The work of \cite{liang2015learning} offers sharp bounds on the excess surrogate risk via localized Rademacher complexity. Finally, \cite{zhu2022efficientactivelearningabstention} introduce a framework for bounding label complexity in terms of the disagreement coefficient. As both the results of \cite{hanneke_2019} and \cite{zhu2022efficientactivelearningabstention} are for surrogate-minimization based active binary classification, we provide a more detailed comparison with their results later in the paper.

Further related is the recent work of \cite{sekhari2023selective}, which develops surrogate-based regression algorithms in the setting of selective sampling—an online variant of active learning in which the data generation process may be arbitrary, even adversarial. Separately, a number of works have investigated nonparametric active learning under smoothness and regularity assumptions \cite{locatelli2017adaptivitynoiseparametersnonparametric, kpotufe2022nuancesmarginconditionsdetermine, minsker2012plugin, hanneke2017nonparametric}.

%% file: section_2/preliminaries.tex
\section{Preliminaries}\label{preliminaries}

\textbf{Regression-Based Classification:}
We consider the problem of surrogate-based multi-class classification in the statistical learning framework, where input instances belong to a set $\calX$ and their labels belong to the finite set $\calY = [K]$ for $K\geq 2$. The learner is given access to a sample $S = \{(x_1,y_1),\ldots,(x_n,y_n)\}$ drawn i.i.d. from an unknown distribution $\calD\in \Delta(\calX\times \calY)$. The learner selects a function $f$ from a class of real-valued functions $\calF \subseteq \{f : \calX \to \bbR^K\}$. A classifier $h_f : \calX \to \calY$ is obtained by applying a fixed decision rule to functions in $\calF$ (e.g., argmax), inducing the hypothesis class $\calH_\calF := \{h_f : f \in \calF\}$.  The learner's objective is to minimize excess classification risk, defined as $$
\Eclass(f, \calF) := \bbE_{(x,y)\sim \calD} [\ind\{h_f(x) \neq y\}] - \inf_{f' \in \calF} \bbE_{(x,y)\sim \calD} [\ind\{h_{f'}(x) \neq y\}].
$$ 
We denote $\Eclass(f):= \Eclass(f, \calF)$ when $\calF$ is the class of all measurable $f$. Since the 0--1 loss is neither convex nor continuous, we instead optimize a convex, differentiable surrogate loss $\ell : \mathbb{R}^K \times \mathcal{Y} \to \mathbb{R}$ over $\mathcal{F}$ minimizing the excess surrogate risk, defined as
$$\mathcal{E}_\ell(f, \mathcal{F}) := \bbE_{(x,y)\sim\calD}[\ell(f(x),y)] - \inf_{f' \in \calF} \bbE_{(x,y)\sim\calD}[\ell(f'(x),y)],$$ 
in hopes it serves as a tractable proxy for the excess classification risk. We also denote $h^* = h_{f^*}$, for $f^*:=\argmin_{f\in\calF} \bbE_{(x,y)\sim \calD}[\ell(f(x),y)]$, to be the classifier induced by the function in $\calF$ that achieves the minimum surrogate risk.

In this work, we consider surrogate loss functions $\ell_\Phi : \mathbb{R}^K \times \mathcal{Y} \to \mathbb{R}$ of the form  
$$
\ell_\Phi(v, y) = \Phi(v) - v[y],
$$  
where $\Phi : \mathbb{R}^K \to \mathbb{R}$ is a $\beta_\Phi$-strongly convex, $L_\Phi$ smooth function in its first argument over the set of realizable score vectors (see \cite{pmlr-v28-agarwal13} for more details). That is, for all $x, x' \in \mathcal{X}$ and all $f \in \mathcal{F}$,  
$$
\frac{\beta_{\Phi}}{2}\,\|f(x) - f(x')\|_2^2 \leq \Phi(f(x))-\Phi(f(x')) - \langle \nabla \Phi(f(x')), f(x)-f(x')\rangle \leq \frac{L_\Phi}{2}\,\|f(x) - f(x')\|_2^2.
$$ 
Each such loss function admits a link function $\phi : \mathbb{R}^K \to \Delta^K$, given by  $\phi(v) = \nabla \Phi(v),$  
which maps the minimizer of the surrogate loss over $\calD$, denoted by $ f_\eta $, to the conditional probability vector $\eta(x) = (\mathbb{P}[Y = c \mid X = x])_{c \in [K]}$.  Moreover, the $\beta_\Phi$-strong convexity of $\Phi$ ensures that $\ell_\Phi$ is strongly convex in its first argument, and the $L_\Phi$ smoothness of $\Phi$ ensures the link function $\phi$ is $L_\Phi$-Lipschitz on the set of realizable score vectors. We then define the regression-based classifier $h_f$, for any $f$, as:
\begin{align*}
    h_f(x) &:= \bfe_{c_f(x)} \,\,\,\text{ where }\,\,\, c_f(x) = \argmax_{c\in K} \phi(f(x))[c],
\end{align*}
ensuring that the Bayes optimal predictor under the surrogate loss induces an optimal classifier, i.e. that $f_{\eta} \in \argmin_{f}\bbE_{(x,y)\sim \calD} [\ind\{h_f(x) \neq y\}].$ Below we instantiate our framework for a couple of common surrogate losses.

\begin{itemize}
    \item \emph{Squared-Loss: }
    If we select $\Phi(v) = \tfrac{1}{2}\|v\|_2^2$, the surrogate loss is equivalent to the squared loss, i.e., $\ell_{sq}(\hat{y},y) = \|\hat{y}-\bfe_y\|_2^2$. In this case, the link function $\phi$ is the identity function, i.e., $\phi(v) = v$. In this case, the Bayes optimal predictor is simply the conditional probability function (i.e., $f_\eta = \eta$).

    \item \emph{Logistic-Loss:}
    If we select $\Phi(v) = \log \big(\sum_{j=1}^{K} e^{v[j]}\big)$, the surrogate loss is equivalent to the logistic loss, i.e., $\ell_{log}(\hat{y},y) = -\log ( \hat{y}[y])$. In this case, the link function is the Boltzmann mapping, i.e., $\phi(v)[i] = e^{v[i]}/\sum_{j=1}^{K} e^{v[j]}$. 
\end{itemize}
We assume access to an  offline regression oracle for the surrogate loss, defined as follows:
\begin{definition}[Offline Regression Oracle]\label{Offline Regression Oracle}
    Given a class of functions $\calF$ an offline regression oracle is specified by mapping $\Oracle : \cup_{t=0}^\infty (\calX \times \calY)^t \mapsto \calF$ and is such that for any distribution $\calD \in \Delta(\calX \times \calY)$ and any $n$, given sample  $S = \{(x_i,y_i)\}_{i\in [n]}$ of $n$ data drawn i.i.d. from this distribution $\calD$, the output of the oracle $\hat{f} =\Oracle(S) $ is such that with probability at least $1-\delta$ over draw of samples, $$\mathcal{E}_{\ell_\Phi}(\hat{f}, \calF) \leq \frac{\comp_{\ell_\Phi}(\calF, \delta, n, K)}{n}.  $$
\end{definition}
\textbf{Learning Protocols:}
We consider two standard statistical learning protocols for multi-class classification: the passive learning protocol and the active learning protocol.

\emph{Passive Learning:} 
In the passive learning setting, the learner receives the labeled sample $S$ drawn i.i.d. from an unknown distribution $\calD $. Given a function class $\calF$ and a surrogate loss $\ell_{\Phi}$ adhering to the specifications above, the learner's objective is to output a function $\hat{f}$ that achieves low excess classification risk $\Eclass(\hat{f})$ as defined earlier.

\emph{Active Learning:}
In the active learning setting, the learner has access to an unlabeled sample $U = \{x_1, \dots, x_n\}$ drawn i.i.d. from $\calD_\calX$, the marginal distribution over $\calX$. The learner may adaptively query the label $y_i$ of any example $x_i \in U$. Then, given access to a function class $\calF$ and a surrogate loss $\ell_{\Phi}$, the learner's objective is to output a function $\hat{f}$ that achieves low classification excess risk $\Eclass(\hat{f})$ while simultaneously minimizing the number of queries $N$ it makes to the labeling oracle.

\textbf{Additional Notation and Definitions:}
For any probability vector $v\in \Delta^K$ and decision $c\in [K]$, we define:
\begin{align*}
    \margin(v) &:= v[\class_v]-\max_{\class\neq \class_v}v[\class'] \, \text{ where } \,\class_v = \argmax_{\class''\in [K]} v[\class'']\\
    \gap(v,\class) &:= \max_{\class'}v[\class']-v[\class].
\end{align*}
Here, $\margin(v)$ quantifies the strength of the decision given by the probability vector $v$ by measuring how much its largest value exceeds the next best alternative. Meanwhile, $\gap(v, \class)$ expresses how much the probability vector favors its top choice over a specific class $\class$, capturing the bias toward the decision relative to any given alternative.

\begin{definition}[Massart Noise Condition, \cite{Massart_2006}] \label{massart noise assumption}
    The marginal distribution $\calD_\calX$ satisfies the Massart noise condition with parameter $\gamma \in [0, \tfrac{1}{2}]$ if $\bbP_{x\sim \calD_\calX}[\margin(\eta(x)) < \gamma] = 0$.
\end{definition}

\begin{definition}[Tsybakov Noise Condition, \cite{tsybakov2004optimal}] \label{tsybakov noise assumption}
    The marginal distribution $\calD_\calX$ satisfies the Tsybakov noise condition with parameter $\beta\geq0$ and a universal constant $c > 0$ if $\bbP_{x\sim \calD_\calX}[\margin(\eta(x)) < \gamma] \leq c\gamma^\beta$ for any $\gamma > 0$.
\end{definition} 

\begin{definition}[Pseudo Dimension, \cite{pollard1984convergence}; \cite{HAUSSLER89, HAUSSLER95}]
Consider a set of real-valued functions $\calF : \calX \to [0,1]$. The pseudo-dimension $\mathrm{Pdim}(\calF)$ of $\calF$ is defined as the VC dimension of the set of threshold functions $\{(x, \zeta) \mapsto \mathbbm{1}(f(x) > \zeta) : f \in \calF\}$.
\end{definition}

%% file: section_3/prelude.tex
\section{Prelude: Passive Binary Classification}

In this section, we provide bounds on excess classification risk in terms of excess surrogate risk, directly using this result to guarantee the performance of our active learning algorithm.  Although we focus on the case of binary classification via squared-error regression, our analysis easily extends to the general learning framework provided in \cref{preliminaries} and we provide the proof in \cref{{bound on multiclass classification excess risk}}.

We can equivalently express the problem of binary classification using a single-coordinate formulation by considering a function class where each function models the probability that the label of $x$ is class 1. Specifically, we redefine the label space as $\calY = \{0,1\}$ and have that $\calF$ consists of functions $f : \calX\to[0,1]$ which, rather than predicting the full conditional probability vector, estimate $\eta(x) = \bbP[Y=1|X=x]$.\footnote{We overload notation for $\eta(x)$. In this special case, it refers to the conditional probability of the label of $x$ being class $1$, but for our more general framework it refers to the conditional probability vector from \cref{preliminaries}} Under this formulation, the regression-based classifier induced by any $f\in \calF$ is given by $h_f(x) = \ind\{f(x)>\tfrac{1}{2}\}$ and the squared loss simplifies to $\ell_{sq}(\hat{y},y) = (\hat{y}-y)^2$.

To obtain meaningful bounds on classification excess risk, we assume that the classifier induced by the best-in-class function agrees with that of the Bayes optimal predictor under squared loss, and that the confidence of the best-in-class function bears a nondecreasing relationship to the confidence of the Bayes optimal predictor.
\begin{assumption}\label{passive binary classification assumption}
    For conditional probability function $\eta(x)$ and function class $\mathcal{F}$:
    \begin{enumerate}
        \item \label{passive binary classification assumption: bullet 1}
        $\bbP_{x\sim \calD_\calX} [h_{f^*}(x) = h_{f_\eta}(x)] = 1$, 
        \item \label{passive binary classification assumption: bullet 2}
        There exists a non-decreasing function $\psi : [0,\tfrac{1}{2}]\to[0,\tfrac{1}{2}]$ such that: 
        $$
            \bbP_{x\sim \calD_\calX} \left[\left|f^*(x)-\tfrac{1}{2}\right| \geq \psi\left(\left|\eta(x)-\tfrac{1}{2}\right|\right)\right] = 1.
        $$
    \end{enumerate}
\end{assumption}

Under this assumption, we are able to prove the following bound on the excess classification risk of a function in terms of its excess surrogate risk.

\begin{proposition}\label{passive binary classification result squared loss}
    For any convex function class $\calF$, if \cref{passive binary classification assumption} holds, then for any $f\in \calF$, 
    \begin{align*}
        \Eclass(f) \leq 2\,\inf_{\gamma}\left\{\Esq(f, \calF)\sup_{a\in (\gamma,1]}\frac{a}{\psi^2\left(a\right)} + \gamma\,\bbP_{x\sim \calD_{\calX}}\left[|\eta(x)-\tfrac{1}{2}|\leq \gamma\right]\right\}
    \end{align*}
\end{proposition}

\begin{proof}
      Starting with a standard analysis of excess risk under \cref{passive binary classification assumption}.\ref{passive binary classification assumption: bullet 1}, we have:
    \begin{align*}
        &\Eclass(f) = \bbE_{(x,y)\sim \calD} [\ind\{h_f(x) \neq y\}] - \bbE_{(x,y)\sim \calD} [\ind\{h_{f_\eta}(x) \neq y\}] = \bbE_{x\sim \calD_{\calX}}\left[\mathbbm{1}\{h_{f}(x)\neq h^*(x)\} \left|2\eta(x)-1\right|\right]
    \intertext{Now, splitting on the event of a $\gamma$ margin on $\eta(x)$ and upper bounding the small-margin case by the maximum margin size times the probability of a data point falling within the margin, we obtain:}
        &\leq 2\,\bbE_{x\sim \calD_{\calX}}\left[\mathbbm{1}\{h_{f}(x)\neq h^*(x)\}\mathbbm{1}\{|\eta(x)-\tfrac{1}{2}|> \gamma\} \left|\eta(x)-\tfrac{1}{2}\right|\right]+2\gamma\,\bbP_{x\sim \calD_{\calX}}\left[|\eta(x)-\tfrac{1}{2}|\leq \gamma\right].
    \end{align*}
    To bound the first expectation, notice that if $h_{f}(x)\neq h^*(x)$ then $|f^*(x)-f(x)| \geq |f^*(x)-\tfrac{1}{2}|$, and by \cref{passive binary classification assumption}.\ref{passive binary classification assumption: bullet 2},  we know $
       |f^*(x)-\tfrac{1}{2}| \geq \psi(|\eta(x)-\tfrac{1}{2}|)$ and thus:
    \begin{align*}
        &\bbE_{x\sim \calD_{\calX}} \left[\mathbbm{1}\{h_{f}(x)\neq h^*(x)\}\mathbbm{1}\{|\eta(x)-\tfrac{1}{2}|> \gamma\} \left|\eta(x)-\tfrac{1}{2}\right|\right] \\
        &\leq \bbE_{x\sim \calD_{\calX}}\left[\mathbbm{1}\{|f^*(x)-f(x)| \geq \psi\left(\left|\eta(x)-\tfrac{1}{2}\right|\right)\}\mathbbm{1}\{|\eta(x)-\tfrac{1}{2}|> \gamma\}\left|\eta(x)-\tfrac{1}{2}\right|\right],
        \intertext{where we can upper bound the first indicator by the ratio of the terms being compared to get:}
        &\leq \bbE_{x\sim \calD_{\calX}}\left[\mathbbm{1}\{|\eta(x)-\tfrac{1}{2}|> \gamma\}\left|\eta(x)-\tfrac{1}{2}\right| \left(\frac{f(x)-f^*(x)}{\psi\left(\left|\eta(x)-\tfrac{1}{2}\right|\right)}\right)^2\right]\\
        &\leq \sup_{a\in (\gamma,1]}\frac{a}{\psi^2\left(a\right)}\bbE_{(x,y)\sim \calD}\left[(f(x)-f^*(x))^2\right]\\
        &\leq \sup_{a\in (\gamma,1]}\frac{a}{\psi^2\left(a\right)}\bbE_{(x,y)\sim \calD}\left[(f(x)-y)^2-(f^*(x)-y)^2\right]\\
        & = \sup_{a\in (\gamma,1]}\frac{a}{\psi^2\left(a\right)} \Esq(f,\calF),
    \end{align*}
    where the final inequality is true by our assumption that $\calF$ is convex. Putting this together with the noise term and optimizing over $\gamma$ gives us our desired result.
\end{proof}

This bound expresses excess classification risk as a balance between the excess squared-error risk and the probability mass near the decision boundary. The infimum over $\gamma$ allows for the tightest tradeoff between these two terms and enables the bound to yield explicit rates when instantiated under margin-based noise models.

%% file: section_4/active_learning.tex
\section{Multi-Class Active Learning via Regression}

We now show it is possible to use our passive learning bounds on classification excess risk to obtain active learning algorithms for classification even for problem instances that are far from realizable.

\textbf{Stream-Based Active Learning Algorithms:}
We study an online variant of active learning, where the learner receives instances sequentially and must decide in real time whether to query each label. In order to minimize the number of label queries, queries are concentrated in regions of uncertainty and therefore the data distribution observed by the learner differs from the underlying distribution. Formally, the learner’s querying strategy is modeled by a function $ q : \calX \to \{0,1\} $, which selects a subset $ Q = \{x \in \calX : q(x) = 1\} $ of inputs to query. The resulting observed distribution $ \calD_Q $ is the renormalization of $ \calD $ restricted to $ Q $:
$$
\calD_Q(x) = 
\begin{cases}
\frac{\calD_\calX(x)}{\bbP_{x' \sim \calD_\calX}[x' \in Q]} & \text{if } x \in Q, \\
0 & \text{otherwise}.
\end{cases}
$$
Notice that a stream-based algorithm can be used for active learning by providing the data points in the unlabeled sample $U$ to the algorithm sequentially.

\textbf{Active Multi-Class Classification Assumption:} 
Since a stream-based active learning algorithm interacts with data drawn from modified distributions $\calD_Q$ determined by its querying strategy, assumptions made solely about the original distribution no longer suffice. We therefore posit a stronger condition, analogous to \cref{passive binary classification assumption}, but required to hold for all modified distributions that may arise during the algorithm's execution:
\begin{assumption}\label{active classification assumption}
There exists a nondecreasing function $\psi : [0,1]\to[0,1]$ such that, for any $Q\subseteq \calX$, if we define $f^*_Q = \argmin_{f\in \calF} \bbE_{(x,y)\sim \calD_Q} \left[ \ell_{\Phi}(f(x),y)\right]
$, then for all $k\in [K]$ we have:
$$
    \bbP_{x\sim \calD_\calX} [\gap(\phi(f^*_{Q}(x)),\class) \geq \psi\left(\gap(\phi(f_\eta(x)),\class)\right)] = 1.
$$
\end{assumption}

\textbf{Active Multi-Class Classification Algorithm:}
Our algorithm is an epoch-based, improper learning algorithm. During each epoch $m\in [M]$, the algorithm employs a consistent query condition $q_{m-1}$ ensuring that all observed data points during the epoch are drawn from the same modified data distribution $\calD_m$. Then, after enough data is observed during the epoch, the learner uses an offline regression oracle, formally defined in \cref{Offline Regression Oracle}, to obtain a good approximation $\hat{f}_m\in \calF$ of the optimal function $f^*_{m}$ in $\calF$ over the modified data distribution, where we denote $f^*_m = f^*_{\calX_m}$ for notational convenience.

Under \cref{active classification assumption}, whenever the classifiers induced by $\hat{f}_m$ and $f^*_m$ agree on a point $x \in \calX_m$, the learner can safely assign a label. To facilitate this, the learner constructs a subset $\calF_m \subseteq \calF$ centered around $\hat{f}_m$, consisting of all functions consistent with the observed data in epoch $m$, such that $f^*_m$ is contained in $\calF_m$ with high probability. Then, the algorithm can confidently predict the label of any point $x\in \calX_m$ on which all classifiers induced by functions in $\calF_m$ agree. Then, the learner updates its query condition such that it only continues to query points $x\in \calX_m$ for which there exists a pair of functions $ f,f' \in \calF_m$ such that $h_f(x) \neq h_{f'}(x)$. 

Finally, the algorithm outputs the classifier $\hat{h}$ which, for any input $x$, considers the smallest epoch $i\in [M]$ for which there did not exist a pair of functions $f,f'\in \calF_i$ such that $h_f(x) \neq h_{f'}(x)$. If such an $i$ exists, it outputs the classification of the consensus; otherwise, it defaults to $h_{\hat{f}_M}(x)$.

As can be seen above, the excess risk and the label complexity of \cref{algorithm1} are directly related to the probability the query function is triggered. This probability can be shown to be bounded via the \emph{Value Function Disagreement Coefficient} defined as follows. 

\input{section_4/algorithm1}

\begin{definition}[Value Function Disagreement Coefficient, \cite{Foster2020}]\label{disagreement coefficient}
    For any $f^* \in \calF$ and $\gamma_0,\epsilon_0>0$, let:
    \begin{align*}
   \theta_{\mathrm{val}}(\calF, \gamma_0, \epsilon_0,f^*) =  \sup_{\calD_\calX} \sup_{\gamma > \gamma_0, \, \varepsilon > \varepsilon_0} \Big\{& \frac{\gamma^2}{\varepsilon^2} \bbP_{\calD_\calX} \Big( \exists f \in \calF : \|f(x) - f^*(x)\|_2 > \gamma, \, \| f - f^* \|_{\calD_\calX} \leq \varepsilon \Big) \Big\} \vee 1,
    \end{align*}
    where $\| f - f^* \|_{\calD_\calX} := \bbE_{x\sim \mathcal{D_X}}[\|f(x)-f^*(x))\|_2^2]$. Also, $\theta_{\mathrm{val}}(\calF, \gamma) = \sup_{f^* \in \calF, \epsilon >0}  \theta_{\mathrm{val}}(\calF, \gamma, \epsilon,f^*)$.
\end{definition}

The following theorem provides a bound on the excess risk and query complexity of\cref{algorithm1}. 

\begin{theorem}\label{active classification result}
    For any convex function class $\calF$, if \cref{active classification assumption} holds, then for the predictor $\hat{f}$ returned by \cref{algorithm1}  using the offline regression oracle in \cref{Offline Regression Oracle} as a subroutine, we have that with probability at least $1-\delta$,
    \begin{align*}
        \Eclass(\hat{f})&\leq \tilde{\calO} \left(\inf_{\gamma >0}\left\{\frac{ L_{\Phi}\beta_\Phi^{-1}\comp_{\ell_{\Phi}}(\calF,\delta,n,K)\log \delta^{-1}}{ n}\hspace{-1mm}\sup_{a\in(\gamma,1]}\frac{a}{\psi^2\left(a\right)}+ \gamma\,\bbP [\margin(\eta(x))\leq\gamma]\right\} \right),
    \end{align*}
    and simultaneously, the number of label queries is bounded as 
    \begin{align*}
        N &\leq  \tilde{\calO}\bigg(\inf_{\gamma >0}\bigg\{\frac{L_\Phi^2\beta_\Phi^{-1} \comp_{\ell_{\Phi}}(\calF,\delta,n,K)\log \delta^{-1}}{\psi^2\left(\gamma\,\right)}\theta_{\mathrm{val}}\left(\calF,\psi\left(\gamma\,\right)\right) +n\bbP\left[\margin(\eta(x))\leq \gamma\right]\bigg\}\bigg),
    \end{align*}
    where $\comp_{\ell_\Phi}(\calF,\delta,n,K)$ is the rate achieved by the offline regression oracle. The $\tilde{\calO}$ hides constants, $\log$ factors in $\comp, \theta_{\mathrm{val}},$ and $n$, and $\log\log$ factors in $\delta$.
\end{theorem}

We instantiate \cref{active classification result} for the Tsybakov noise model for settings where \cref{active classification assumption} holds for $\psi(x)=x$, an assumption far weaker than realizability, and provide bounds on labeled and unlabeled sample complexity in terms of the error rate. We provide a proof in \cref{tsybakov noise proofs}.
\begin{corollary}\label{binary classification squared loss corollary}
    For any convex function class $\calF$, if \cref{active classification assumption} holds for $\psi(x)=x$ and Tsybakov's noise condition for parameter $\beta\geq 0$, then for the predictor $\hat{f}$ returned by \cref{algorithm1} using the offline regression oracle in \cref{Offline Regression Oracle} as a subroutine, we have that with probability at least $1-\delta$,
    \begin{align*}
        n \leq \tilde{\calO}\left(\comp_{\ell_{\Phi}}(\calF,\delta,n,K)\epsilon^{-\frac{\beta+2}{\beta+1}}\log\delta^{-1}\right),
    \end{align*}
    and,
    \begin{align*}
        N \leq \tilde{\calO}\left(\comp_{\ell_{\Phi}}(\calF,\delta,n,K)\theta_{\mathrm{val}}^{\frac{\beta}{\beta+2}} \epsilon^{-\frac{2}{\beta+1}}\log\delta^{-1}\right).
    \end{align*}
    The $\tilde{\calO}$ hides constants, $\log$ factors in $\comp, \theta_{\mathrm{val}},$ and $\epsilon$, and $\log\log$ factors in $\delta$.
\end{corollary}

\textbf{Our Algorithm vs. Prior Methods:} 
\cref{table1} compares the rates achieved by \cref{algorithm1} to those of state-of-the-art surrogate minimization-based active learning algorithms under the widely studied Tsybakov noise model. Notably, our sample complexity, like that of \cite{hanneke_2019}, is independent of the disagreement coefficient, and our label complexity exhibits a strictly better dependence on it than both \cite{hanneke_2019} and \cite{zhu2022efficientactivelearningabstention}.\footnote{Note that the analysis in \cite{hanneke_2019} is expressed in terms of the \emph{sign-based disagreement coefficient}, denoted $\theta_{\mathrm{sgn}}$, which was first introduced in \cite{hanneke2007bound}.} Furthermore, the dependence on the target error rate $\epsilon$ in both the sample and label complexity of our algorithm matches that of \cite{hanneke_2019} and \cite{zhu2022efficientactivelearningabstention}.\footnote{Although our guarantees are not expressed in terms of the Pseudo-dimension, it upper bounds the rate achieved by the ERM in the binary classification setting with squared loss. See \cref{table discussion} for more details.}

\begin{table}[h]
\centering
\begin{tabular}{@{} l p{0.21\textwidth} l l @{}}
\toprule
Algorithm & Assumption & Sample Complexity ($n$) & Label Complexity ($N$) \\
\midrule
\cite{hanneke_2019} & Realizability & $\tilde{\calO}\left(d\epsilon^{-\frac{\beta+2}{\beta+1}}\log\delta^{-1}\right)$ & $\tilde{\calO}\left(d\theta_{\mathrm{sgn}} \epsilon^{-\frac{2}{\beta+1}}\log\delta^{-1}\right)$ \\
\cite{zhu2022efficientactivelearningabstention} & Realizability & $\tilde{\calO}\left(d\theta_{\mathrm{val}}\epsilon^{-\frac{\beta+2}{\beta+1}}\log\delta^{-1}\right)$ & $\tilde{\calO}\left(d\theta_{\mathrm{val}}\epsilon^{-\frac{2}{\beta+1}}\log\delta^{-1}\right)$ \\
\cref{algorithm1} & \cref{active classification assumption} where $\psi(x)=x$, convex $\calF$
& $\tilde{\calO}\left(d\epsilon^{-\frac{\beta+2}{\beta+1}}\log\delta^{-1}\right)$ 
& $\tilde{\calO}\left(d\theta_{\mathrm{val}}^{\frac{\beta}{\beta+2}} \epsilon^{-\frac{2}{\beta+1}}\log\delta^{-1}\right)$ \\
\bottomrule
\end{tabular}
\caption{Comparison of sample and label complexities of squared-error regression based active classification algorithms. In this table $d=\mathrm{PDim}(\calF)$ and the $\tilde{\calO}$ hides constants, $\log$ factors in $d, \theta,$ and $\epsilon$, and $\log\log$ factors in $\delta$.}
\label{table1}
\end{table}

\textbf{Our Assumptions vs. Realizability:}
While it is evident that our assumptions are weaker than realizability, one might ask whether they remains qualitatively weaker than approximate versions of the realizability condition. In the binary case, when paired with Massart noise, \cref{passive binary classification assumption} is implied under $\mathcal{L}_\infty(\calD_\calX)$ $\epsilon$-realizability (i.e., when $|f^*(x)-\eta(x)| \leq \epsilon$ for all $x\in \calX$). We formalize this claim below and provide a proof in \cref{our assumptions vs realizability}.
\begin{claim}\label{massarts and pointwise claim}
    For $\gamma \geq \epsilon >0$, if a squared-error regression problem instance is both $\mathcal{L}_\infty(\calD_\calX)$ $\epsilon$-realizable and satisfies \cref{massart noise assumption} for parameter $\gamma$ then, \cref{passive binary classification assumption} holds for the given problem and data distribution with $\psi(x) = (1-\tfrac{\epsilon}{\gamma})x$. 
\end{claim}
In fact, even when the problem satisfies Tsybakov's noise condition and is only $\mathcal{L}_2(\calD_\calX)$ $\epsilon$-approximately realizable—that is, when $\bbE_{x \sim \calD_\calX} \left[(f^*(x) - \eta(x))^2\right] \leq \epsilon$—one can still obtain a version of \cref{passive binary classification assumption} which holds outside a region of small probability measure. While \cref{passive binary classification result squared loss} is shown under the original condition which is stated to hold almost surely, our analysis naturally extends to the setting where \cref{passive binary classification assumption} is violated on a subset of the input space with measure at most a function of $\epsilon$ and we would pay this measure additively in our bounds.

Furthermore, there exist instances where the problem is far from approximately realizable (i.e., where $\epsilon$ is a constant), yet \cref{active classification assumption}—and consequently, our guarantees—still hold. The following example illustrates such a setting, see \cref{our assumptions vs realizability} for more details.
\begin{example}\label{not approximately realizable example}
        Consider $\calX = \{\pm \vec{e}_i : i \in [d]\}$ and $\calF = \{x \mapsto \tfrac{1+w\cdot x}{2} : \|w\|_2 \leq 1\}$. Let $\calD_\calX = \Uniform[\calX]$, and $\eta(x) = \tfrac{1 + \vec{1}\cdot x}{2}$. For this example, \cref{active classification assumption} holds with $\psi(x) = d^{-1/2}x$. That is, for all $Q\subseteq \calX$ that could be induced by \cref{algorithm1}, for all $x\in Q$ we have:
        \begin{align*}
            |f^*_{Q}(x)-\tfrac{1}{2}| \geq \psi(|\eta(x)-\tfrac{1}{2}|).
        \end{align*}
        However, we also have that:
        $$\bbE_{x \sim \calD_\calX} \left[(f^*(x) - \eta(x))^2\right] = \left(\tfrac{1}{2} - \tfrac{1}{2}d^{-1/2}\right)^2,$$ 
        i.e., the instance is not $\mathcal{L}_2(\calD_\calX)\ \epsilon$-realizable for any $\epsilon < \left(\tfrac{1}{2} - \tfrac{1}{2}d^{-1/2}\right)^2$.
\end{example}

The above claim and example should convince the reader that our assumption is provably weaker than approximate versions of realizability and can be far from them. In fact, our assumption intuitively says that the optimal regression function is only not allowed to be close to the decision boundary in regions where the true label is fairly decisive. It does not preclude cases when the regression function is very confident in places where the true label is close to the margin. This is a natural requirement, as the primary source of excess classification error is when the true label has a clear bias that the regression solution is unable to capture.

\textbf{The Hurdle of Non-Realizability:}
Previous works in surrogate-based active learning algorithms \cite{hanneke_2019, zhu2022efficientactivelearningabstention} have all made the realizability assumption. Under realizability, in any epoch $m$, the minimizer of the surrogate loss in $\calF$ on $\calD_{m}$ is always the Bayes optimal predictor, which would be contained in $\calF$. This consistency allowed for the learner to progressively aggregate data across epochs and refine a version space that, with high probability, contains the Bayes optimal predictor. The final classifier would then be chosen from $\calF$ in direct alignment with this version space. 

In contrast, our algorithm is fundamentally \emph{improper}. Since we make no realizability assumption, the surrogate loss minimizer within~$\calF$ may not be the Bayes optimal predictor and can vary arbitrarily across epochs. This invalidates the version space construction central to prior approaches. Instead, to ensure robust performance under our new assumption, our algorithm treats each epoch independently and learns an approximation $\hat{f}_m \in \calF$ to the local surrogate-optimal function $f^*_{m}$ using only the labeled data observed during that epoch. The final classifier stitches together the epoch-wise approximations, assigning to each point the prediction of the earliest approximation that can be deemed correct with high confidence.

\textbf{The Importance of Convexity of $\calF$:}
\cref{active classification assumption} alone is not sufficient to guarantee the success of our passive or active learning results, as demonstrated by the following simple example:
\begin{example}
Let $\gamma > 0$, $\delta > 0$, and define $\calX = \{0\}$ with $\eta(0) = \tfrac{1}{2} + \gamma$. Let $\calF = \{f, f^*\}$ where $f^*(0) = \tfrac{1}{2} + 2\gamma$ and $f(0) = \tfrac{1}{2} - \delta$. Then $\Eclass(f,\calF)=2\gamma$, while $\Esq(f, \calF) \to 0$ as $\delta \to 0$. 
\end{example}
This example illustrates that without additional structural assumptions on $\calF$, excess classification risk may not be controlled by excess surrogate risk. We can reconcile this example by additionally requiring that there exists a constant $C>0$ for which: 
$$
    \bbE_{x\sim \calD_\calX} [(f(x)-f^*(x))^2] \leq C\mathcal{E}_{\ell_\Phi}(f, \calF).
$$
This is exactly the role convexity plays in the proof of \cref{passive binary classification result squared loss} (see \cref{convexity discussion} for  further discussion). We specifically posit convexity of $\calF$ as it is both the most tangible structural assumption under which this relationship provably holds and a standard one in the literature on surrogate loss minimization (e.g., \cite{bartlett2006convexity, hanneke_2019}). Furthermore, convex function classes include many widely used, expressive model families—such as linear predictors, generalized linear models, and kernel-based methods (e.g., RKHSs)—and are well-suited for optimization-based learning algorithms commonly employed in modern practice.

While we focus on convex $\calF$, we note that our analysis lends itself to the weaker condition of star convexity around each optimal predictor $f^*_m$; however, such conditions are significantly more difficult to verify in practice.

%% file: section_4/algorithm1.tex
\begin{algorithm}[hbt!]\onehalfspacing
    \caption{\label{algorithm1}\text{Active Learning in Epochs}}
    \begin{algorithmic}[1]
    \State \textbf{Parameters:}  $\delta\in (0,1)$
    \State Define $\tau_m = 2^m-1, \tau_0=0$, and $q_0(x)=1$ and $B :=  C \log^3(n) \comp_{\ell_\Phi}(\calF, \delta, n)$.
    \For{$m = 1,\ldots, M$}
        \For{$t = \tau_{m-1}+1,\ldots, \tau_m$}
            \State Receive $x_t$ for $(x_t,y_t) \sim \mathcal{D}$
            \If{$q_{m-1}(x_t) = 1$}
                \State Query the label $y_t$ of $x_t$
            \EndIf
        \EndFor
        \State Compute estimate $\hat{f}_{m} \leftarrow \mathsf{Alg}^{\mathsf{OR}}_{\ell_\Phi}(S_m,\calF)$ for $S_m = \{(x_t,y_t) : q_{m-1}(x_t) = 1, t \in [\tau_{m-1}+1,\tau_m]\}$:
        \State Implicitly Construct Set
        $\mathcal{F}_m :=
             \left\{f\in \mathcal{F} : \sum_{t=\tau_{m-1}+1}^{\tau_{m}}q_{m-1}(x_t)\|f(x_t)-\hat{f}_m(x_t)\|_2^2\leq B \right\}$
        \State Update condition $q_{m}(x) \gets q_{m-1}(x)\cdot\mathbbm{1}\left\{\exists f,f' \in \calF_m \,\,s.t.\,\, h_f(x) \neq h_{f'}(x)\right\}$
    \EndFor
    \State \Return $\hat{f}$ defined by:
    $$
    \hat{f}(x) := 
        \begin{cases}
            \hat{f}_i(x) & \text{if $q_M(x)=0$ and $i$ is the smallest index s.t. $\nexists f,f' \in \calF_i,  h_f(x) \neq h_{f'}(x)$,} \\
            \hat{f}_M(x) & \text{otherwise.} 
        \end{cases}
    $$
    \end{algorithmic}
\end{algorithm}

%% file: section_5/discussion.tex
\section{Discussion}
Our results show that in the batch setting, for both the active and passive learning, even under relaxations of the stringent realizability assumption commonly made in regression-based classification literature, one can obtain effectively the same guarantees as proven under realizability. We note that a compelling part of our analysis of the performance of \cref{algorithm1} directly incorporates our passive learning result as a black-box component. This shows that in general, bounds on excess classification risk in terms of excess surrogate risk under relaxations of realizability appear to be a gateway to relaxing realizability in other learning paradigms as well. We conclude with a discussion of avenues for future work:

\textbf{Algorithms for Interactive Learning:} Regression-based algorithms for the contextual bandit problem under realizability are provided in \cite{foster2020beyond} for the worst case, and for instance dependent bounds, in \cite{Foster2020}. One can ask the question of whether this realizability assumption can be relaxed with milder assumptions like the one we mention above for the multi-class setting. We do note however that since, in these reductions, one picks distributions over actions in an epoch using the current estimate of $\hat{f}_{m}(x)$, the modified distribution over context-action pairs is not only a function of the benchmark class, but also this estimate. One can still change the assumption to having modified data distributions over $\calX$ space but any distribution over actions. However, we are yet to carefully analyze the implications of such an assumption. 

\textbf{Exploring the Additional Power of Being Improper:} We note that our active learning algorithm is improper and technically can work better than proper passive learning algorithms in certain scenarios. While we have toy examples that illustrate this, further principled investigation into when this happens could be interesting.

%% file: section_6/appendix.tex
\input{section_6/multiclass_proofs}
\input{section_6/other_proofs}

%% file: section_6/multiclass_proofs.tex
\section{Proof of Main Result (\cref{active classification result})}

We restate \cref{active classification result} for the reader's convenience. In the proof of this theorem, we denote the number of points whose label was queried during epoch $m$ as $k_m$.

\begin{theorem}[Theorem 4.2 Restated] \label{active classification result (appendix)}
    For any convex function class $\calF$, if \cref{active classification assumption} holds, then for the predictor $\hat{f}$ returned by \cref{algorithm1}  using the offline regression oracle in \cref{Offline Regression Oracle} as a subroutine, we have that with probability at least $1-\delta$,
    \begin{align*}
        \Eclass(\hat{f})&\leq \tilde{\calO} \left(\inf_{\gamma >0}\left\{\frac{ L_{\Phi}\beta_\Phi^{-1}\comp_{\ell_{\Phi}}(\calF,\delta,n,K)\log \delta^{-1}}{ n}\hspace{-1mm}\sup_{a\in(\gamma,1]}\frac{a}{\psi^2\left(a\right)}+ \gamma\,\bbP [\margin(\eta(x))\leq\gamma]\right\} \right),
    \end{align*}
    and simultaneously, the number of label queries is bounded as 
    \begin{align*}
        N &\leq  \tilde{\calO}\bigg(\inf_{\gamma >0}\bigg\{\frac{L_\Phi^2 \beta_\Phi^{-1}\comp_{\ell_{\Phi}}(\calF,\delta,n,K)\log \delta^{-1}}{\psi^2\left(\gamma\,\right)}\theta_{\mathrm{val}}\left(\calF,\psi\left(\gamma\,\right)\right) +n\bbP\left[\margin(\eta(x))\leq \gamma\right]\bigg\}\bigg),
    \end{align*}
    where $\comp_{\ell_\Phi}(\calF,\delta,n,K)$ is the rate achieved by the offline regression oracle. The $\tilde{\calO}$ hides constants, $\log$ factors in $\comp, \theta_{\mathrm{val}},$ and $n$, and $\log\log$ factors in $\delta$. 
\end{theorem}

\begin{proof}
    We start by bounding the excess risk of the classifier outputted by \cref{algorithm1} and then bound the number of queries it makes. Starting with the reformulation of excess risk from \cref{excess risk gap reformulation}, we have,
    \begin{align*}
        \Eclass(h_{\hat{f}}) &= \bbE_{x\sim \calD_\calX} [\ind\{h_{\hat{f}}(x)\neq h_{f_\eta}(x)\}\gap(\eta(x), \class_{\hat{f}}(x))],
        \intertext{where by decomposing on the query condition, we get:}
        &= \bbE_{x\sim \calD_\calX} [\ind\{q_{M}(x) = 0, h_{\hat{f}}(x)\neq h_{f_\eta}(x)\}\gap(\eta(x), \class_{\hat{f}}(x))]\\
        &\quad+\bbE_{x\sim \calD_\calX} [\ind\{q_{M}(x) = 1, h_{\hat{f}}(x)\neq h_{f_\eta}(x)\}\gap(\eta(x), \class_{\hat{f}}(x))].
    \end{align*}
    We will now separately bound the excess risk incurred when the classifier would have chosen not to query, i.e. when $q_{M}(x) = 0$,  and when it would have chosen to query, i.e. when $q_{M}(x) = 1$, under the intersection of the high probability events of \cref{expected distance to subdistribution minimizers look like average empirical distance to them} and \cref{number of queries in epochs looks like their expectation}.
    
    We begin by bounding the excess risk incurred when the classifier would have chosen not to query. For any $x\in \calX$ such that $q_{M}(x) = 0$, from the definition of \cref{algorithm1}, we have that $h_{\hat{f}}(x) = h_{\hat{f}_{i}}(x)$, where $i$ is the earliest epoch such that every function $f\in \calF_i$ is in consensus on $x$. By \cref{subdistribution minimizers are in version spaces}, we know that $f^*_{i}\in \calF_{i}$ and therefore that $h_{f^*_{i}}(x) = h_{\hat{f}_{i}}(x) = h_{\hat{f}}(x)$. Finally, by \cref{active classification assumption}, we know that $h_{f^*_{i}}(x) =h_{f_\eta}(x)$, implying that the classifier does not incur risk when it would not have queried.
    
    We now bound the excess risk incurred when the classifier would have chosen to query. First note that this can be bounded by the probability that the algorithm queries a data point it encounters in the final epoch:
    \begin{align*}
        \bbE_{x\sim \calD_\calX} [\ind\{q_{M}(x) = 1, h_{f}(x)\neq h_{f_\eta}(x)\}\gap(\eta(x), \class_f(x))] \leq \bbP_{x\sim \calD_{\calX}} [q_M(x)=1] \leq \bbP_{x\sim \calD_{\calX}} [q_{M-1}(x)=1].
    \end{align*}
    Accordingly, in the case this probability is small, less than $\tfrac{4}{n_M} \log \left(\tfrac{M}{\delta}\right)$, then so is the algorithm's excess risk. Therefore, we are left to consider the case when this probability is at least $\tfrac{4}{n_M} \log \left(\tfrac{M}{\delta}\right)$. With this in mind, excess risk can be rewritten as:
    \begin{align*}
        &\bbE_{x\sim \calD_{\calX}}\left[\ind\{q_{M}(x) = 1, h_{\hat{f}}(x)\neq h_{f_\eta}(x)\}\gap(\eta(x), \class_f(x))\right]\\
        &= \bbE_{x\sim \calD_{\calX}}\left[\ind\{h_{\hat{f}}(x)\neq h_{f_\eta}(x)\}\gap(\eta(x), \class_f(x))\Big\vert q_{M}(x) = 1\right] \bbP_{x\sim \calD_\calX}[q_{M}(x) = 1]\\
        &= \bbE_{x\sim \calD_{\calX_m}}\left[\ind\{h_{\hat{f}}(x)\neq h_{f_\eta}(x)\}\gap(\eta(x), \class_f(x))\right] \bbP_{x\sim \calD_\calX}[q_{M}(x) = 1],
        \intertext{where, by \cref{active classification assumption}, we have:}
        &= \bbE_{x\sim \calD_{\calX_M}}\left[\ind\{h_{\hat{f}}(x)\neq h_{f^*_m}(x)\}\gap(\eta(x), \class_f(x))\right] \bbP_{x\sim \calD_\calX}[q_{M}(x) = 1].
    \end{align*}
    Now, notice, the expectation term is simply the excess risk of the function $\hat{f}$ as seen in \cref{excess risk gap reformulation}. Therefore, we can invoke \cref{passive multiclass classification result} giving us that, for any $\gamma>0$:
    \begin{align*}
        &\bbE_{x\sim \calD_{\calX_M}}\left[\ind\{h_{\hat{f}}(x)\neq h_{f^*_M}(x)\}\gap(\eta(x), \class_f(x))\right] \bbP_{x\sim \calD_\calX}[q_{M}(x) = 1] \\
        &\leq \left(8\,L_\Phi\beta_{\Phi}^{-1}\frac{\comp(\calF,\delta,k_m)}{k_m}\sup_{a\in (\gamma,1]}\frac{a}{\psi^2\left(a\right)}+\gamma\,\bbP_{x\sim \calD_\calX} [\margin(\eta(x))\leq\gamma]\right) \bbP_{x\sim \calD_\calX}[q_{M}(x) = 1]\\
        &\leq 8\,L_\Phi\beta_{\Phi}^{-1}\frac{\comp(\calF,\delta,k_m)}{k_m}\sup_{a\in(\gamma,1]}\frac{a}{\psi^2\left(a\right)}\bbP_{x\sim \calD_\calX}[q_{M}(x) = 1] + \gamma\,\bbP_{x\sim \calD_\calX} [\margin(\eta(x))\leq\gamma]\\
        \intertext{where, since $q_m(x) \leq q_{m-1}(x)$ by construction, we have:}
        &\leq 8\,L_\Phi\beta_{\Phi}^{-1}\frac{\comp(\calF,\delta,k_m)}{k_m} \sup_{a\in(\gamma,1]}\frac{a}{\psi^2\left(a\right)}\bbP_{x\sim \calD_\calX}[q_{M-1}(x) = 1] + \gamma\,\bbP_{x\sim \calD_\calX} [\margin(\eta(x))\leq\gamma]\\
        &\leq 8\,L_\Phi\beta_{\Phi}^{-1}\frac{\comp(\calF,\delta,k_m)}{k_m}\frac{\bbE[k_m]}{n_M} \sup_{a\in(\gamma,1]}\frac{a}{\psi^2\left(a\right)} + \gamma\,\bbP_{x\sim \calD_\calX} [\margin(\eta(x))\leq\gamma]\\
        \intertext{Then, by the lower bound from \cref{number of queries in epochs looks like their expectation}, we have that:}
        &\leq 8\,L_\Phi\beta_{\Phi}^{-1}\frac{\comp(\calF,\delta,k_m)}{\frac{1}{2}\,\bbE[k_m]-\log (\tfrac{M}{\delta})} \frac{\bbE[k_m]}{n_M}\sup_{a\in(\gamma,1]}\frac{a}{\psi^2\left(a\right)} + \gamma\,\bbP_{x\sim \calD_\calX} [\margin(\eta(x))\leq\gamma],\\
        \intertext{where since, $\bbP_{x\sim \calD_{\calX}} [q_{M-1}(x)=1]\geq \tfrac{4}{n_M} \log \left(\tfrac{M}{\delta}\right)\implies \bbE[k_m] \geq 4 \log \left(\tfrac{M}{\delta}\right)$, and therefore:}
        &\leq 8\,L_\Phi\beta_{\Phi}^{-1}\frac{\comp(\calF,\delta,k_m)}{\frac{1}{4}\,\bbE[k_m]} \frac{\bbE[k_m]}{n_M}\sup_{a\in(\gamma,1]}\frac{a}{\psi^2\left(a\right)} + \gamma\,\bbP_{x\sim \calD_\calX} [\margin(\eta(x))\leq\gamma]\\
        \intertext{Now, because $\comp$ is increasing in its third argument, we obtain that:}
        &\leq 32\, L_\Phi\beta_{\Phi}^{-1}\frac{\comp(\calF,\delta,n,K)}{n_M}\sup_{a\in(\gamma,1]}\frac{a}{\psi^2\left(a\right)} + \gamma\,\bbP_{x\sim \calD_\calX} [\margin(\eta(x))\leq\gamma]\\
        &= 64\, L_\Phi\beta_{\Phi}^{-1}\frac{\comp(\calF,\delta,n,K)}{n}\sup_{a\in(\gamma,1]}\frac{a}{\psi^2\left(a\right)}+ \gamma\,\bbP_{x\sim \calD_\calX} [\margin(\eta(x))\leq\gamma],
    \end{align*}
    where in the last line we plug in $M=\log n$, and by extension $n_M=\tfrac{n}{2}$. Doing the same substitution for the other case and taking the maximum of the two recovers our claimed bound on excess risk.

    We now bound the number of queries made by \cref{algorithm1}. We know by \cref{number of queries in epochs looks like their expectation} that,
    \begin{align*}
        N &= \sum_{m=1}^{M} k_m \\
        &\leq \sum_{m=1}^{M} \left(\frac{3}{2}\, \bbE[k_m]+\log \frac{M}{\delta} \right) \\
        &= \sum_{m=1}^{M} \frac{3}{2}\, n_m \bbP_{x\sim \calD_{\calX}} [q_{m-1}(x) = 1]+\log \frac{M}{\delta}.
        \intertext{Where, by plugging in the upper bound for $\bbP_{x\sim \calD_{\calX}} [q_{M}(x) = 1]$ we get by \cref{probability of query conditioning firing} , we get:}
        &\leq \sum_{m=1}^{M} \frac{3}{2}\, n_m   \frac{4}{n_{m-1}}\bigg(\max \bigg\{\frac{L_\Phi^2(9\,C(\calF,\delta,n,K)+32\,\beta_\Phi^{-1}\comp(\calF,\delta,n,K))}{ \psi^2\left(\gamma\,\right)}\theta_{\mathrm{val}}\left(\calF, \psi\left(\gamma\,\right)\right),\log \frac{M}{\delta}\bigg\}\bigg)\\
        &\quad+\frac{3}{2}\, n_m \bbP_{x\sim D_{\calX}}\left[\margin(\eta(x))\leq \gamma\right]+\log \frac{M}{\delta}\\
        &\leq 12\,\log n\bigg(\max \bigg\{\frac{L_\Phi^2(9\,C(\calF,\delta,n,K)+32\,\beta_\Phi^{-1}\comp(\calF,\delta,n,K))}{ \psi^2\left(\gamma\,\right)}\theta_{\mathrm{val}}\left(\calF, \psi\left(\gamma\,\right)\right),\log \frac{\log n}{\delta}\bigg\}\bigg)\\
        &\quad +\frac{3}{2}\,n\bbP_{x\sim D_{\calX}}\left[\margin(\eta(x))\leq \gamma\right]+\log \frac{\log n}{\delta},
    \end{align*}
    where in the last line we plug in $M=\log n$ and recognize that for any parameter setting $C(\calF,\delta,n,K) = \tilde{\calO}(\comp_{\ell_\Phi}(\calF,\delta,n,K))$ to recover our claimed bound on label complexity.
\end{proof}

\subsection{Bound on Multi-Class Classification Excess Risk}\label{bound on multiclass classification excess risk}

We generalize \cref{passive binary classification result squared loss} to the regression-based multi-class classification framework presented in \cref{preliminaries}. This result is used as a blackbox component of our analysis of the labeled and unlabeled complexity of \cref{algorithm1}. Under the following assumption, we are able to prove the result below.

\begin{assumption}\label{passive multiclass classification assumption}
    There exists a non-decreasing function $\psi : (\gamma,1]\to[0,1]$ such that, for all $\class\in [K]$, we have: 
    $$
        \bbP_{x\sim \calD_\calX} [\gap(\phi(f^*(x)),\class) \geq \psi\left(\gap(\phi(f_\eta(x)),\class)\right)] = 1.
    $$
\end{assumption}

\begin{proposition}\label{passive multiclass classification result}
    For any convex function class $\calF$, if \cref{passive multiclass classification assumption} holds, then for any $f\in \calF$, 
    \begin{align*}
        \Eclass(f) \leq \inf_{\gamma}\left\{4\,L_\Phi\beta_{\Phi}^{-1}\mathcal{E}_{\ell_{\Phi}}(f,\calF)\sup_{a\in (\gamma,1]}\frac{a}{\psi^2\left(a\right)}+\gamma\,\bbP_{x\sim \calD_\calX} [\margin(\eta(x))\leq\gamma]\right\}.
    \end{align*}
\end{proposition}

\begin{proof}
    Starting with the definition of classification excess risk, we have:
    \begin{align}
        \Eclass(f) &= \bbE_{(x,y)\sim \calD} [\ind\{h_f(x) \neq y\}] - \bbE_{(x,y)\sim \calD} [\ind\{h_{f_\eta}(x) \neq y\}]\nonumber\\
        &= \bbE_{(x,y)\sim \calD} [\ind\{h_{f}(x)\neq h_{f_\eta}(x)\}(\ind\{h_{f_\eta}(x)= y\}-\ind\{h_{f}(x)= y\})] \nonumber\\
        &= \bbE_{x\sim \calD_\calX} [\ind\{h_{f}(x)\neq h_{f_\eta}(x)\}\gap(\eta(x), \class_f(x))].\label{excess risk gap reformulation}
        \intertext{Then, splitting on the event that $\eta(x)$ has a $\gamma$-gap with respect to the label $\class_f(x)$, we have:}
        &= \bbE_{x\sim \calD_\calX} [\ind\{h_{f}(x)\neq h_{f_\eta}(x)\}\gap(\eta(x), \class_f(x))\ind\{\gap(\eta(x), \class_f(x))>\gamma\}]\nonumber\\
        &\quad+\bbE_{x\sim \calD_\calX} [\ind\{h_{f}(x)\neq h_{f_\eta}(x)\}\gap(\eta(x), \class_f(x))\ind\{\gap(\eta(x), \class_f(x))\leq\gamma\}],\nonumber\\
        &= \bbE_{x\sim \calD_\calX} [\ind\{h_{f}(x)\neq h_{f_\eta}(x)\}\gap(\eta(x), \class_f(x))\ind\{\gap(\eta(x), \class_f(x))>\gamma\}]\nonumber\\
        &\quad+\gamma\bbE_{x\sim \calD_\calX} [\ind\{h_{f}(x)\neq h_{f_\eta}(x)\}\ind\{\gap(\eta(x), \class_f(x))\leq\gamma\}],\nonumber
        \intertext{where, since $h_{f}(x)\neq h_{f_\eta}(x)$, we know $\gap(\eta(x), \class_f(x)) \geq \margin(\eta(x))$, and therefore:}
        &= \bbE_{x\sim \calD_\calX} [\ind\{h_{f}(x)\neq h_{f_\eta}(x)\}\gap(\eta(x), \class_f(x))\ind\{\gap(\eta(x), \class_f(x))>\gamma\}]\nonumber\\
        &\quad+\gamma\bbP_{x\sim \calD_\calX} [\margin(\eta(x))\leq\gamma].\nonumber
    \end{align}
    It remains to bound the first term. To do this, recall from \cref{passive multiclass classification assumption} that $\psi\left(\gap(\eta(x),\class_f(x))\right)\leq \gap(\phi(f^*(x)),\class_f(x))$. Then, since: 
    \begin{align*}
        \gap(\phi(f^*(x)),\class_f(x)) &\leq \gap(\phi(f^*(x)),\class_f(x)) + \gap(\phi(f(x)),\class_{f^*}(x))\\
        &\leq 2\,\|\phi(f^*(x))-\phi(f(x))\|_\infty \\
        &\leq 2\,\|\phi(f^*(x))-\phi(f(x))\|_2,
    \end{align*}
    we have that:
    \begin{align*}
        &\bbE_{x\sim \calD_\calX} [\ind\{h_{f}(x)\neq h_{f_\eta}(x)\}\gap(\eta(x), \class_f(x))\ind\{\gap(\eta(x), \class_f(x))>\gamma\}]\\
        &\leq \bbE_{x\sim \calD_\calX} [\ind\{2\,\|\phi(f^*(x))-\phi(f(x))\|_2\geq \psi\left(\gap(\eta(x),\class_f(x))\right)\}\gap(\eta(x), \class_f(x))\ind\{\gap(\eta(x), \class_f(x))>\gamma\}]
        \intertext{where, since the ratio of the terms being compared in the first indicator is an upper bound on the indicator, we have:}
        &\leq \bbE_{x\sim \calD_\calX} \left[\gap(\eta(x), \class_f(x))\ind\{\gap(\eta(x), \class_f(x))>\gamma\}\left(\frac{2\,\|\phi(f^*(x))-\phi(f(x))\|_2}{\psi\left(\gap(\eta(x),\class_f(x))\right)}\right)^2\right]\\
        &\leq 4\sup_{a\in (\gamma,1]}\frac{a}{\psi^2\left(a\right)}\bbE_{x\sim \calD_\calX} \left[\|\phi(f^*(x))-\phi(f(x))\|_2^2\right]\\
        \intertext{where, since $\phi$ is $L_\Phi$-Lipschitz over the set of realizable score vectors, we have:}
        &\leq 4\,L_\Phi\sup_{a\in (\gamma,1]}\frac{a}{\psi^2\left(a\right)}\bbE_{x\sim \calD_\calX} \left[\|f^*(x)-f(x)\|_2^2\right]\\
        &\leq 8\,L_\Phi\beta_{\Phi}^{-1}\sup_{a\in (\gamma,1]}\frac{a}{\psi^2\left(a\right)}\mathcal{E}_{\ell_{\Phi}}(f,\calF)
    \end{align*}
    where the final inequality is because $\calF$ is convex. Putting this together with the noise term and optimizing over $\gamma$ gives us our desired result.
\end{proof}

\subsection{Concentration Lemmas and Supporting Results}

In the proof of our main result we bound the unlabeled and labeled sample complexities of \cref{algorithm1} under a good event. In this section, we show that this good event happens with high probability. This good event is the event that for ever epoch $m\in [M]$:
\begin{enumerate}
    \item the number of queries $k_m$ made within the epoch is concentrated around its expectation, and
    \item the empirical distance from any function $f\in \calF$ to $f^*_m$ on labeled data observed in epoch $m$ is concentrated around its $L_2(\calD_{\calX_m})$ distance to $f^*_m$.
\end{enumerate}
We independently show that each of these events happen with high probability – in \cref{number of queries in epochs looks like their expectation} and \cref{expected distance to subdistribution minimizers look like average empirical distance to them} respectively – and therefore the intersection of these events also happens with high probability. 

\begin{lemma}\label{number of queries in epochs looks like their expectation}
    For all $m\in [M]$, with probability $1-2\delta$, $$\frac{1}{2}\bbE[k_m] -\log \frac{M}{\delta}\leq k_m \leq \frac{3}{2}\bbE[k_m]+\log \frac{M}{\delta}.$$
\end{lemma}

\begin{proof}
We provide a proof of the lower bound using the lower tail Chernoff bound. First, note that $k_m$ is a Binomial random variable, as it is a sum of i.i.d. Bernoulli random variables each representing whether the learner queried on a round of epoch $m$. By the lower tail Chernoff bound for a sum of independent Bernoulli random variables, we have that for any $\epsilon\in (0,1)$,
$$\Pr\left[k_m < (1-\epsilon)\bbE[k_m]\right] \leq \exp\left(-\frac{\epsilon^2 \bbE[k_m]}{2}\right).$$
Setting the probability of this bad event to be $\delta/M$, and union bounding over $m\in [M]$ gives us that with probability at least $1-\delta$, for all $m\in [M], k_m < (1-\epsilon)\bbE[k_m]$ for $\epsilon = \sqrt{\tfrac{-2\log(\delta/M)}{\bbE[k_m]}}$. Now, by the AM-GM inequality, we have,
$$\epsilon = \sqrt{\frac{-2\log(\delta/M)}{\bbE[k_m]}} \leq \frac{1}{2} + \frac{\log(M/\delta)}{\bbE[k_m]},$$
where plugging this upper bound in for $\epsilon$ gives us our desired lower bound. The upper bound follows identically from using the upper tail Chernoff bound, also with probability at least $1-\delta$, implying they happen simultaneously with probability at least $1-2\delta$.
\end{proof}

\begin{lemma}
\label{expected distance to subdistribution minimizers look like average empirical distance to them}
    Fix any $\delta\in (0,1)$. Then, for all $m\in [M]$ and any $f\in \calF$, with probability $1-\delta$, we have:
    \begin{align*}
        &\bbE_{x\sim \calD_{\calX_m}}\left[\|f(x)-f^*_m(x)\|_2^2\right] \leq 2\left(\frac{1}{k_m}\sum_{t=\tau_{m-1}+1}^{\tau_{m}} \ind\{q_{m-1}(x_t) =1\}\|f(x_t)-f^*_{m}(x_t)\|_2^2\right)+\frac{C(\calF,\delta,n, K)}{k_m},
    \end{align*}
    and
    \begin{align*}
        &\frac{1}{k_m}\sum_{t=\tau_{m-1}+1}^{\tau_{m}} \ind\{q_{m-1}(x_t)=1\}\|f(x_t)-f^*_{m}(x_t)\|_2^2  \leq 2\,\bbE_{x\sim \calD_{\calX_m}} [\|f(x)-f^*_m(x)\|_2^2]+\frac{C(\calF,\delta,n, K)}{k_m},
    \end{align*}
    where $C(\calF,\delta,n, K) = C'\left(Kn\log^3(n)\mathrm{Rad}^2_n(\calF)+\log\left(\tfrac{K\log n}{\delta}\right)\right)$ for some absolute constant $C'$.
\end{lemma}

\begin{proof}
    Take $m\in [M]$. Recall the definition of $f^*_m$ to be the best-in-class function on the subdistribution induced by our query condition in epoch $m$. Then, for any $f\in \calF$, consider the average squared distance between $f$ and $f^*_m$ over labeled data observed during epoch $m$. Since, each data point $(x_t,y_t)$ of this epoch is sampled from $\calD$, and its label $y_t$ is observed exactly when $q_{m-1}(x_t)=1$, we can imagine each data point whose label was observed as being sampled from $\calD_m$ --- the original data distribution $\calD$ normalized after being restricted to the set $\calX_{m}$. Therefore, if we denote the rounds for which \cref{algorithm1} did query in epoch $m$ as $t^m_{1},\ldots,t^m_{k_m}$ we have:
    \begin{align*}
        \frac{1}{k_m}\sum_{t=\tau_{m-1}+1}^{\tau_{m}} \ind\{q_{m-1}(x_t)=1\}\|f(x_t)-f^*_{m}(x_t)\|_2^2 &= \frac{1}{k_m}\sum_{t=\tau_{m-1}+1}^{\tau_{m}} \ind\{x_t\in \calX_m\}\|f(x_t)-f^*_{m}(x_t)\|_2^2 \\
        &= \frac{1}{k_m}\,\sum_{i=1}^{k_m} \,\|f(x_{t^m_i})-f^*_{m}(x_{t^m_i})\|_2^2.
    \end{align*}
    Then, by applying the upper and lower bounds on this quantity from \cref{concentration around minimizer of distribution}, union bounding over all $m\in [M]$, and re-normalizing $\delta$, we get our desired lemma except with the additive term $C(\calF,\delta,k_m, K)$. Finally, we remark that since $C(\calF,\delta,k_m, K)$ is an increasing function in its third argument we can replace $k_m$ with $n$.
\end{proof}

\begin{lemma}
\label{concentration around minimizer of distribution}
    Let $\calD\in \Delta(\calX\times \calY)$ , $\calF \subseteq \{f \mid f : \calX \to [0,1]^K\}$ , $n\geq 2$, and $\delta\in (0,1)$. Then with probability at least $1-\delta$ over samples $S = \{(x_i,y_i)\}_{i\in [n]}$ drawn i.i.d. from $\calD$, the following inequalities hold for all $f,f'\in \calF$:
    \begin{align*}
        \bbE_{x\sim \calD_{\calX}} [\|f(x) - f'(x)\|^2_2] \leq 2\left(\frac{1}{n}\sum_{i=1}^n [\|f(x_i) - f'(x_i)\|^2_2]\right)+\frac{C(\calF,\delta,n, K)}{n},
    \end{align*}
    and,
    \begin{align*}
        \frac{1}{n}\sum_{i=1}^n \|f(x_i)-f'(x_i)\|_2^2 \leq 2\,\bbE_{x\sim \calD_{\calX}} [\|f(x_i)-f'(x_i)\|_2^2]+ \frac{C(\calF,\delta,n, K)}{n},
    \end{align*}
    for $C(\calF,\delta,n, K) = C'\left(Kn\log^3(n)\mathrm{Rad}^2_n(\calF)+\log\left(\tfrac{K\log n}{\delta}\right)\right)$ some absolute constant $C'$.
\end{lemma}
\begin{proof}
We directly apply the result from \cref{empirical distance between pairs of functions are concentrated around the true distance} and extend to the multi-class setting. In particular, we consider $\calF^k := \{f(\cdot)[k]:f \in \calF\}$ to be the class of $k$-th coordinate functions corresponds to the functions in $\calF$. Now, the guarantees from \cref{concentration around minimizer of distribution} hold for all $\calF^k$ for each $k\in[K]$. Taking a union bound over all $K$ coordinates gives our result. 
\end{proof}

\begin{lemma}[{\cite{1444, Rakhlin_2017}}] 

\label{empirical distance between pairs of functions are concentrated around the true distance}
    Let $\calD\in \Delta(\calX\times \calY)$ , $\calF \subseteq \{f \mid f : \calX \to [0,1]\}$ , $n\geq 2$, and $\delta\in (0,1)$. Then with probability at least $1-\delta$ over samples $S = \{(x_i,y_i)\}_{i\in [n]}$ drawn i.i.d. from $\calD$, the following inequalities hold for all $f,f'\in \calF$:
    \begin{align*}
        \bbE_{x\sim \calD_{\calX}} [(f(x)-f'(x))^2] \leq 2\left(\frac{1}{n}\sum_{i=1}^n (f(x_i)-f'(x_i))^2\right)+\frac{C(\calF,\delta,n)}{n},
    \end{align*}
    and,
    \begin{align*}
        \frac{1}{n}\sum_{i=1}^n (f(x_i)-f'(x_i))^2 \leq 2\,\bbE_{x\sim \calD_{\calX}} [(f(x)-f'(x))^2]+ \frac{C(\calF,\delta,n)}{n},
    \end{align*}
    for $C(\calF,\delta,n) = C'\left(n\log^3(n)\mathrm{Rad}^2_n(\calF)+\log\left(\tfrac{\log n}{\delta}\right)\right)$ some absolute constant $C'$.
\end{lemma}

\subsection{Bounding the Probability of Querying Via the Disagreement Coefficient}

We show in our main result that the labeled sample complexity can be bounded given a bound on the probability that our query condition is triggered. The following lemma bounds this probability for any epoch in terms of the value-based disagreement coefficient.

\begin{lemma}\label{probability of query conditioning firing}
    Under the high probability event of \cref{number of queries in epochs looks like their expectation}, for all $m \in [M]$, we have:
    \begin{align*}
        \bbP_{x\sim \calD_\calX}[q_{m}(x) = 1] &\leq  \frac{4}{n_m}\left(\max \left\{\frac{L_\Phi^2(9\,C(\calF,\delta,n,K)+32\,\beta_\Phi^{-1}\comp(\calF,\delta,n,K))}{ \psi^2\left(\gamma\,\right)}\theta_{\mathrm{val}}\left(\calF, \psi\left(\gamma\,\right)\right), \log \frac{M}{\delta}\right\}\right)\\
        &\quad+\bbP_{x\sim D_{\calX}}\left[\margin(\eta(x))\leq \gamma\right].
    \end{align*}
\end{lemma}

\begin{proof}
    We consider two cases and independently prove a bound on our desired quantity for each case. Then, the final bound is the max over the two cases. 
    
    \emph{Case 1:} If $\bbE[k_{m}]\leq 4 \log \left(\tfrac{M}{\delta}\right)$, then we have,
    \begin{align*}
        \bbE[k_{m}] = n_{m} \bbP_{x \sim \calD_{\calX}}[q_{m-1}(x) = 1] \leq 4 \log \left(\tfrac{M}{\delta}\right)
    \end{align*}
    which implies that,
    \begin{align*}
        \bbP_{x \sim \calD_{\calX}}[q_{m}(x) = 1] \leq \bbP_{x \sim \calD_{\calX}}[q_{m-1}(x) = 1] \leq \frac{4 \log \left(\tfrac{M}{\delta}\right)}{n_{m}}.
    \end{align*}
    
    \emph{Case 2:} Otherwise, if $\bbE[k_{m}]> 4 \log \left(\tfrac{M}{\delta}\right)$, then by the construction of \cref{algorithm1}, we can decompose the query condition on epoch $m+1$ as,
    \begin{align*}    
        \bbP_{x\sim \calD_\calX}[q_{m}(x) = 1] &= \bbP_{x\sim D_{\calX}}\left[\exists f,f' \in \calF_m \,\,s.t.\,\, h_f(x) \neq h_{f'}(x),\, q_{{m-1}}(x)=1\right]\\ 
        &= \bbP_{x\sim D_{\calX}}\left[\exists f,f' \in \calF_m \,\,s.t.\,\, h_f(x) \neq h_{f'}(x),\, q_{{m-1}}(x)=1, \margin(\eta(x))> \gamma\right]\\ 
        &\quad+\bbP_{x\sim D_{\calX}}\left[\exists f,f' \in \calF_m \,\,s.t.\,\, h_f(x) \neq h_{f'}(x),\, q_{{m-1}}(x)=1, \margin(\eta(x))\leq \gamma\right],
        \intertext{where we can upper bound the second term by the probability $x\sim \calD$ falls inside the margin to get:}
        &\leq \bbP_{x\sim D_{\calX}}\left[\exists f,f' \in \calF_m \,\,s.t.\,\, h_f(x) \neq h_{f'}(x),\, q_{{m-1}}(x)=1, \margin(\eta(x))> \gamma\right]\\
        &\quad+\bbP_{x\sim D_{\calX}}\left[\margin(\eta(x))\leq \gamma\right].
    \end{align*}
    Now, to bound the first term in this sum, we rewrite it as the following conditional probability:
    \begin{align*}
        &\bbP_{x\sim D_{\calX}}\left[\exists f,f' \in \calF_m \,\,s.t.\,\, h_f(x) \neq h_{f'}(x),\, q_{{m-1}}(x)=1, \margin(\eta(x))> \gamma\right]\\
        &= \bbP_{x\sim D_{\calX}}\left[\exists f,f' \in \calF_m \,\,s.t.\,\, h_f(x) \neq h_{f'}(x),\, \margin(\eta(x))> \gamma \biggm\vert q_{m-1}(x)=1\right]\bbP_{x\sim \calD_{\calX}} [q_{m-1}(x)=1]\\
        &= \bbP_{x\sim D_{\calX}}\left[\exists f,f' \in \calF_m \,\,s.t.\,\, h_f(x) \neq h_{f'}(x),\, \margin(\eta(x))> \gamma \biggm\vert x\in \calX_{m}\right]\bbP_{x\sim \calD_{\calX}} [q_{m-1}(x)=1]\\
        &= \bbP_{x\sim D_{\calX_{m}}}\left[\exists f,f' \in \calF_m \,\,s.t.\,\, h_f(x) \neq h_{f'}(x),\, \margin(\eta(x))> \gamma\right]\bbP_{x\sim \calD_{\calX}} [q_{m-1}(x)=1].
    \end{align*}
    To bound the first term in this product, we start by recalling that from \cref{subdistribution minimizers are in version spaces}, we know $f^*_{m}\in \calF_m$ for all $m\in [M]$. Then, for any $x\in \calX_{m}$ for which $\exists f,f' \in \calF_m$ such that $h_f(x) \neq h_{f'}(x)$, there must exist a function $f\in \calF_m$ for which $\|f^*_{m}(x)-f(x)\|_2 \geq \tfrac{1}{L_\Phi}\|\phi(f^*_{m}(x))-\phi(f(x))\|_2\geq \tfrac{1}{L_\Phi}\margin(\phi(f^*_m(x)))$. Furthermore, we know by \cref{active classification assumption} that $\margin(\phi(f^*_m(x)))> \psi(\margin(\eta(x)))$. However, since $f$ is in $\calF_m$, by \cref{subdistribution minimizer is close to all functions in expectation} we also know an upper bound on $\|f-f^*_m\|^2_{\calD_{\calX_{m}}}$. Therefore, we can bound by the probability these two events happen simultaneously, to get:
    \begin{align*}
        &\bbP_{x\sim D_{\calX_{m}}}\left[\exists f,f' \in \calF_m \,\,s.t.\,\, h_f(x) \neq h_{f'}(x),\, \margin(\eta(x))> \gamma\right]\\
        &\leq \bbP_{x\sim D_{\calX_{m}}}\bigg[\margin(\eta(x))> \gamma, \exists f\in \calF_m : \|f(x)-f^*_{m}(x)\|_2 > \frac{\psi\left(\margin(\eta(x))\right)}{L_\Phi},\\
        &\hspace{20mm} \|f-f^*_{m}\|^2_{\calD_{\calX_{m}}}\leq \frac{9\,C(\calF,\delta,n,K)+32\,\beta_\Phi^{-1}\comp(\calF,\delta,n,K)}{k_m} \bigg].
        \intertext{Then, since $\psi$ is a non-decreasing function, we have:}
        &\leq \bbP_{x\sim D_{\calX_m}}\bigg[\exists f\in \calF_m : \|f(x)-f^*_{m}(x)\|_2 > \frac{\psi\left(\gamma\,\right)}{L_\Phi}, \\
        &\hspace{20mm}\|f-f^*_{m}\|^2_{\calD_{\calX_{m}}} \leq \frac{9\,C(\calF,\delta,n,K)+32\,\beta_\Phi^{-1}\comp(\calF,\delta,n,K)}{k_m} \bigg]
        \intertext{where by the definition of the disagreement coefficient, we have:}
        &\leq \frac{L_\Phi^2}{\psi^2\left(\gamma\,\right)}\frac{9\,C(\calF,\delta,n,K)+32\,\beta_\Phi^{-1}\comp(\calF,\delta,n,K)}{k_m}\\
        &\quad\cdot\theta_{\mathrm{val}}\bigg(\calF_m, \psi\left(\gamma\,\right),\frac{9\,C(\calF,\delta,n,K)+32\,\beta_\Phi^{-1}\comp(\calF,\delta,n,K)}{k_m},f^*_{m}\bigg)\\
        &\leq \frac{L_\Phi^2}{\psi^2\left(\gamma\,\right)}\frac{9\,C(\calF,\delta,n,K)+32\,\beta_\Phi^{-1}\comp(\calF,\delta,n,K)}{k_m}\theta_{\mathrm{val}}\left(\calF, \psi\left(\gamma\,\right)\right).
    \end{align*}
    Now, from \cref{number of queries in epochs looks like their expectation}, we know that, with probability at least \(1 - \delta\), the following inequality holds:
    $$ k_m \geq  \frac{1}{2}\, \bbE[k_m]-\log \left(\tfrac{M}{\delta}\right) = \frac{1}{2}\, n_m \bbP_{x \sim \calD_{\calX}}[q_{m-1}(x) = 1]-\log \left(\tfrac{M}{\delta}\right).$$
    From this lower bound and our assumption, we have that,
    \begin{align*}
        &\frac{L_\Phi^2}{\psi^2\left(\gamma\,\right)}\frac{9\,C(\calF,\delta,n,K)+32\,\beta_\Phi^{-1}\comp(\calF,\delta,n,K)}{k_m}\theta_{\mathrm{val}}\left(\calF, \psi\left(\gamma\,\right)\right)\\
        &\leq \frac{4L_\Phi^2}{\psi^2\left(\gamma\,\right)}\frac{9\,C(\calF,\delta,n,K)+32\,\beta_\Phi^{-1}\comp(\calF,\delta,n,K)}{n_m  \bbP_{x \sim \calD_{\calX}}[q_{m-1}(x) = 1]}\theta_{\mathrm{val}}\left(\calF, \psi\left(\gamma\,\right)\right).
    \end{align*}
    Plugging this upper bound back in and simplifying gives us,
    \begin{align*}
        &\bbP_{x\sim D_{\calX}}\left[\exists f,f' \in \calF_m \,\,s.t.\,\, h_f(x) \neq h_{f'}(x),\, q_{{m-1}}(x)=1, \margin(\eta(x))> \gamma\right]+\bbP_{x\sim D_{\calX}}\left[\margin(\eta(x))\leq \gamma\right] \\
        &\leq \frac{4L_\Phi^2}{\psi^2\left(\gamma\,\right)}\frac{9\,C(\calF,\delta,n,K)+32\,\beta_\Phi^{-1}\comp(\calF,\delta,n,K)}{n_m}\theta_{\mathrm{val}}\left(\calF, \psi\left(\gamma\,\right)\right)+\bbP_{x\sim D_{\calX}}\left[\left|\eta(x)-\tfrac{1}{2}\right|\leq \gamma\right].
    \end{align*}
    Finally, by taking the max of the two bounds from both cases gives us our desired result. 
\end{proof}

\subsection{Supporting Lemmas}

In our main result, we show that bounding excess risk is contingent on showing that the version space $\calF_m$ contains the surrogate minimizer of the modified distribution $f^*_m$. Similarly, our bound on labeled sample complexity is also contingent on this being true. We now prove this holds true under our good event.

\begin{lemma}\label{subdistribution minimizers are in version spaces}
    Under the high probability event of \cref{expected distance to subdistribution minimizers look like average empirical distance to them}, with probability $1-\delta$, it is true that for any $m\in [M]$, $f^*_{m}\in \calF_{m}$.
\end{lemma}

\begin{proof}
    First recall that we denote $t_{i}^m$ to be the $i$-th queried point in epoch $m$. Then, we can bound the empirical distance between $f^*_m$ and $\hat{f}_m$ on queried points in the $m$-th epoch by \cref{expected distance to subdistribution minimizers look like average empirical distance to them}, to get:
    \begin{align*}
        \sum_{t = \tau_{m-1}+1}^{\tau_m} \ind\{q_{m-1}(x_t)=1\}\|\hat{f}_{m}(x_t)-f^*_{m}(x_t)\|_2^2
        &= \sum_{i=1}^{k_m} \|\hat{f}_{m}(x_{t^m_i})-f^*_{m}(x_{t^m_i})\|_2^2\\
        &\leq  2k_m\,\bbE_{x\sim \calD_{\calX_{m}}} [\|f^*_m(x)-\hat{f}_{m}(x)\|_2^2]+C(\calF,\delta,n,K).
    \end{align*}
    Then, since $\calF$ is convex, we have:
    \begin{align*}
        2k_m\,\bbE_{x\sim \calD_{\calX_{m}}} [\|f^*_m(x)-\hat{f}_{m}(x)\|_2^2]+C(\calF,\delta,n, K) &\leq 4\,k_m\beta_\Phi^{-1}\mathcal{E}_{\ell_{\Phi}}(\hat{f}_m,\calF)+C(\calF,\delta,n,K),
        \intertext{where since this is just the excess risk of $\hat{f}_m$, we can apply the bound on the excess risk of the multi-regression oracle to get:}
        &\leq 4\,\beta_\Phi^{-1}\comp_{\ell_\Phi}(\calF,\delta,k_m,K)+C(\calF,\delta,n,K)\\
        &\leq 4\,\beta_\Phi^{-1}\comp_{\ell_\Phi}(\calF,\delta,n,K)+C(\calF,\delta,n,K),
    \end{align*}
    where the final inequality is possible since $\comp_{\ell_\Phi}$ is an increasing function in its third argument. Plugging this back in gives us our desired result.
\end{proof}

Our bound on labeled sample complexity is also contingent on the expected $L_2(\calD_{\calX_m})$ distance between $\hat{f}_m$ and $f^*_m$ being small. We prove this also holds true under our good event.

\begin{lemma}\label{subdistribution minimizer is close to all functions in expectation}
    Under the high probability event of \cref{expected distance to subdistribution minimizers look like average empirical distance to them}, for any $m\in [M]$ and $f\in \calF_{m}$, we have, $$\bbE_ {x\sim \calD_{\calX_{m}}} \left[\|f(x)-f^*_{m}(x)\|_2^2\right] \leq \frac{9\,C(\calF,\delta,n,K)+32\,\beta_\Phi^{-1}\comp_{\ell_{\Phi}}(\calF,\delta,n,K)}{k_m}.$$
\end{lemma}

\begin{proof}
    By \cref{expected distance to subdistribution minimizers look like average empirical distance to them}, we have,
    \begin{align*}
        \bbE_{x\sim \calD_{\calX_m}}\left[\|f(x)-f^*_m(x)\|_2^2\right]
        &\leq 2\,\left(\frac{1}{k_m}\sum_{t=\tau_{m-1}+1}^{\tau_{m}} \ind\{q_m(x_t)=1\}\|f(x_t)-f^*_{m}(x_t)\|_2^2 \right)+\frac{C(\calF,\delta,n,K)}{k_m}.
    \end{align*}
    To bound the summation term, we apply a basic triangle inequality, $\|a-b\|_2^2\leq 2\|a-c\|_2^2+2\|b-c\|_2^2$, 
    \begin{align*}
        &\sum_{t=\tau_{m-1}+1}^{\tau_{m}} \ind\{q_m(x_t)=1\}\|f(x_t)-f^*_{m}(x_t)\|_2^2 \\
        &\leq \sum_{t=\tau_{m-1}+1}^{\tau_{m}} \ind\{q_m(x_t)=1\}\left(2\|f(x_t)-\hat{f}_{m}(x)\|_2^2+2\|\hat{f}_{m}(x)-f^*_{m}(x_t)\|_2^2\right)\\
        &\leq 4\, \sup_{f'\in \calF_{m}}\sum_{t=\tau_{m-1}+1}^{\tau_{m}} \ind\{q_m(x_t)=1\}\|f'(x_t)-\hat{f}_{m}(x)\|_2^2,
        \intertext{where, by the construction of \cref{algorithm1}, we have a bound on the distance from any function in $\calF_m$ to $\hat{f}_m$,}
        &\leq 32\,\beta_\Phi^{-1}\comp_{\ell_\Phi}(\calF,\delta,n,K)+4\,C(\calF,\delta,n,K).
    \end{align*}
    Finally, by plugging this bound back in, we achieve our desired result.
\end{proof}

%% file: section_6/other_proofs.tex
\section{Instantiating for Tsybakov's Noise Condition}

In this section we prove \cref{binary classification squared loss corollary} and briefly discuss how we arrive at our results in Table 1.

\subsection{Proof of \cref{binary classification squared loss corollary}}\label{tsybakov noise proofs}

We restate \cref{binary classification squared loss corollary} for the reader's convenience. Note this is for a generic loss fitting the specifications from \cref{preliminaries}.
\begin{corollary}[Corollary 1 Restated]\label{binary classification squared loss corollary (appendix)}
        For any convex function class $\calF$, if \cref{active classification assumption} holds for $\psi(x)=x$ and Tsybakov's noise condition for parameter $\beta\geq 0$, then for the predictor $\hat{f}$ returned by \cref{algorithm1}  using the offline regression oracle in \cref{Offline Regression Oracle} as a subroutine, we have that with probability at least $1-\delta$,
    \begin{align*}
        n \leq \tilde{\calO}\left(\comp_{\ell_{\Phi}}(\calF,\delta,n,K)\epsilon^{-\frac{\beta+2}{\beta+1}}\log\delta^{-1}\right),
    \end{align*}
    and,
    \begin{align*}
        N \leq \tilde{\calO}\left(\comp_{\ell_{\Phi}}(\calF,\delta,n,K)\theta_{\mathrm{val}}^{\frac{\beta}{\beta+2}} \epsilon^{-\frac{2}{\beta+1}}\log\delta^{-1}\right).
    \end{align*}
    The $\tilde{\calO}$ hides constants, $\log$ factors in $\comp, \theta_{\mathrm{val}},$ and $\epsilon$, and $\log\log$ factors in $\delta$.
\end{corollary}

\begin{proof}
    Instantiating the bound from \cref{active classification result} with $\psi(x)=x$ and Tsybakov's Noise Condition gives us: 
    \begin{align*}
    \Eclass(\hat{f}) = \frac{32\, \beta_\Phi^{-1} L_{\Phi}\comp(\calF,\delta,n,K)}{\gamma n}+ c\gamma^{\beta+1}+\frac{2}{n}\log \frac{\log n}{\delta},
    \end{align*}
    where we purposefully take the tighter bound that appears in the proof of \cref{active classification result}.

    Now, optimizing with respect to $\gamma$ and plugging the optimal value back into our excess risk bound gives
    \begin{align*}
        \Eclass(\hat{f})&\leq \tilde{\calO} \left(\left(\frac{\comp_{\ell_{\Phi}}(\calF,\delta,n,K)}{n}\right)^{\frac{\beta+1}{\beta+2}}+\frac{1}{n}\log \delta^{-1}\right)\\
        &\leq \tilde{\calO} \left(\left(\frac{\comp_{\ell_{\Phi}}(\calF,\delta,n,K)\log \delta^{-1}}{n}\right)^{\frac{\beta+1}{\beta+2}}\right).
    \end{align*}
    Now, upper bounding by the error rate $\epsilon$ and isolating $n$ gives the first part of our result,
    $$n \leq \tilde{\calO}\left(\comp_{\ell_{\Phi}}(\calF,\delta,n,K)\epsilon^{-\frac{\beta+2}{\beta+1}}\log\delta^{-1}\right).$$
    To compute the bound on labeled sample complexity, we again begin by plugging $\psi(x)=x$ and Tsybakov's Noise Condition into the label complexity bound from \cref{active classification result (appendix)}. We again use the tighter bound that appears in the analysis.
    \begin{align*}
    N &\leq  12\,\log n\bigg(\max \bigg\{\frac{L_\Phi^2(9\,C(\calF,\delta,n,K)+32\,\beta_\Phi^{-1}\comp(\calF,\delta,n,K))}{ \gamma^2}\theta_{\mathrm{val}}\left(\calF, \psi\left(\gamma\,\right)\right), \log \frac{\log n}{\delta}\bigg\}\bigg)\\
    &\quad+\frac{3c}{2}\,n\gamma^{\beta}+\log \frac{\log n}{\delta}\\
    &\leq  \tilde{\calO}\left(\frac{\comp(\calF,\delta,n,K)\theta_{\mathrm{val}}\left(\calF\right)}{ \gamma^2}+\log \delta^{-1}+n\gamma^{\beta}\right)
     \end{align*}
    Optimizing over $\gamma$ and plugging the optimal value back into our label complexity bound gives us:
    \begin{align*}
        N &\leq  \tilde{\calO}\left(\left(\comp(\calF,\delta,n,K)\theta_{\mathrm{val}}\left(\calF\right)\right)^{\frac{\beta}{\beta+2}}n^{\frac{2}{\beta+2}}+\log \delta^{-1}\right),
     \end{align*}
     where plugging the $n$ from above back in gives us our desired result.
\end{proof}

\subsection{Discussion of Results in \cref{table1}}\label{table discussion}

We arrive at our results in \cref{table1} for binary classification with squared error by instantiating \cref{binary classification squared loss corollary} for squared-error and bounding the excess risk of the offline regression oracle using the following bound for the ERM. Prior work shows that this quantity can be bounded in terms of the Covering Number defined below.
\begin{definition}[Covering Number] $V$ is an $\ell_2$ cover of $\calF$ on $x_1, \ldots, x_n$ at scale $\beta$ if for all $f \in \calF$, there exists a collection of its elements such that the union of the $\beta$-balls with centers at the elements contains $\calF$. That is, there exists $\mathbf{v}_f \in V$ such that $$\left( \frac{1}{n} \sum_{i = 1}^{n} \left|f(x_i) - \mathbf{v}_f[i]\right| \right)^\frac{1}{2} \leq \beta.$$ The empirical covering number $\calN_2(\calF, \beta; x_1, \ldots, x_n)$ is the size of the minimal set of such a $V$, and we define the covering number as 
    \begin{align*}
        \calN_2(\calF, \beta, n) = \sup_{x_1, \ldots, x_n} \calN_2(\calF, \beta; x_1, \ldots, x_n)
    \end{align*}
\end{definition}
In particular, we have the following bound.
\begin{lemma}\label{Rademacher Complexity Integration}[\cite{liang2015learning}] For any convex function class $\calF$ with probability $1-\delta$, if $\hat{f}$ is the ERM, then
\begin{align*}
    \comp_{sq}(\calF, \delta, n) = n\log^3 n\log\tfrac{1}{\delta} \left(
\inf_{\kappa>0,\,\nu \in [0, \kappa]} \left( 4\nu + \frac{12}{\sqrt{n}} \int_{\nu}^{\kappa} \sqrt{\log \calN_2(\calF, \beta)} \, d\beta \right) + \frac{\log \calN_2(\calF, \kappa) + \log\tfrac{1}{\delta}}{n}\right)
\end{align*}
for  some absolute constant $C_1 > 0$. 
\end{lemma}

\begin{proof}
This is a direct consequence of applying Lemma 7 to the upper bound of Theorem 4 of \cite{liang2015learning}, then upper bounding the resulting complexity by multiplying by an additional $\log^3 n$.
\end{proof}

In key cases, the complexity measure from \cref{Rademacher Complexity Integration} simplifies:  
\begin{itemize}
    \item \emph{Finite pseudo-dimension:}  
    If $\mathcal{F}$ is convex with finite pseudo-dimension (i.e. $\Pdim < \infty$), then \cite{lee1998importance} shows that $\comp_{sq}(\mathcal{F},\delta,n) = \mathcal{O}(\Pdim \log n).$
    
    \item\emph{Bounded Covering Number:}  
    If $\mathcal{F}$ is convex and the covering number at scale $\kappa$ satisfies the following $\log \mathcal{N}_2(\mathcal{F}, \kappa) \leq \kappa^{-p}$ for some $p>0$, then \cite{liang2015learning} shows that: $\comp_{sq}(\mathcal{F},\delta,n) = \calO(n^{p/(2+p)})$ when $p \in (0,2)$, and $\comp_{sq}(\mathcal{F},\delta,n) = n^{1 - 1/p}$ when $p \geq 2$ with an extra logarithmic factor when $p = 2$.
\end{itemize}

\section{Comparing Our Assumptions to Approximate Realizability}\label{our assumptions vs realizability}

In this section, we prove our \cref{massarts and pointwise claim} stating that \cref{passive binary classification assumption} is implied under approximate realizability and Massart's Noise condition and provide a proof that our \cref{not approximately realizable example} is a problem instance satisfying our assumption that is in fact far from being approximately realizable.

\subsection{Proof of \cref{massarts and pointwise claim}}

We restate the \cref{massarts and pointwise claim} for the reader's convenience.

\begin{claim}[Claim 1 Restated]\label{massarts and pointwise claim (appendix)}
    For $\gamma \geq \epsilon >0$, if a squared-error regression problem instance is both $\mathcal{L}_\infty(\calD_\calX)$ $\epsilon$-realizable and satisfies \cref{massart noise assumption} for parameter $\gamma$ then, \cref{passive binary classification assumption} holds for the given problem and data distribution with $\psi(x) = (1-\tfrac{\epsilon}{\gamma})x$. 
\end{claim}

\begin{proof}
To show \cref{passive binary classification assumption}.\ref{passive binary classification assumption: bullet 1} holds, assume for contradiction that there exists $x$ such that $h_{f^*}(x) \neq h_{f_\eta}(x)$. Then, it must be that $|f^*(x) - \eta(x)|\geq \gamma \geq \epsilon$, implying the problem is not $L_\infty(\calD_\calX)$ $\epsilon$-realizable.

Now for \cref{passive binary classification assumption}.\ref{passive binary classification assumption: bullet 2}, starting with the margin on $f^*$, we have by the triangle inequality that:
\begin{align*}
    \left|f^*(x) - \tfrac{1}{2}\right| = \left|f^*(x) - \eta(x) + \eta(x) - \tfrac{1}{2}\right| \geq \left|\eta(x) - \tfrac{1}{2}\right| - \left|f^*(x) - \eta(x)\right| \geq \left|\eta(x) - \tfrac{1}{2}\right| - \epsilon,
\end{align*}
where the final inequality is true since the problem is point-wise $\epsilon$-realizable. Now, consider the event where $|\eta(x) - \tfrac{1}{2}| \geq \gamma$. If we factor out the bias of $\eta$, then under this event, which takes place with probability $1$, we have:
\begin{align*}
    \left|\eta(x) - \tfrac{1}{2}\right| - \epsilon = \left|\eta(x) - \tfrac{1}{2}\right|\left(1-\epsilon\,\left|\eta(x) - \tfrac{1}{2}\right|^{-1}\right) \geq \left|\eta(x) - \tfrac{1}{2}\right|\left(1-\tfrac{\epsilon}{\gamma}\right),
\end{align*} 
and therefore, \cref{passive binary classification assumption}.\ref{passive binary classification assumption: bullet 2} holds with $\psi(x) = (1-\tfrac{\epsilon}{\gamma})x$.
\end{proof}

\subsection{Proof of Correctness for \cref{not approximately realizable example}}

We restate \cref{not approximately realizable example} for the reader's convenience.

\begin{example}[Example 1 Restated]
        Consider $\calX = \{\pm \vec{e}_i : i \in [d]\}$ and $\calF = \{x \mapsto \tfrac{1+w\cdot x}{2} : \|w\|_2 \leq 1\}$. Let $\calD_\calX = \Uniform[\calX]$, and $\eta(x) = \tfrac{1 + \vec{1}\cdot x}{2}$. For this example, \cref{active classification assumption} holds with $\psi(x) = d^{-1/2}x$. That is, for all $Q\subseteq \calX$ that could be induced by \cref{algorithm1}, for all $x\in Q$ we have:
        \begin{align*}
            |f^*_{Q}(x)-\tfrac{1}{2}| \geq \psi(|\eta(x)-\tfrac{1}{2}|).
        \end{align*}
        However, we also have that:
        $$\bbE_{x \sim \calD_\calX} \left[(f^*(x) - \eta(x))^2\right] = \left(\tfrac{1}{2} - \tfrac{1}{2}d^{-1/2}\right)^2,$$ 
        i.e., the instance is not $\mathcal{L}_2(\calD_\calX)\ \epsilon$-realizable for any $\epsilon < \left(\tfrac{1}{2} - \tfrac{1}{2}d^{-1/2}\right)^2$.
\end{example}

\begin{proof}
    First, we claim that the only possible regions of uncertainty that can be induced by the algorithm are of the form $\{\pm e_i : i\in S\}$ for some $S \subseteq [d]$. To see this, note that for any $f\in \calF$, $f(x) = -f(-x)$ and thus $h_{f}(x) \neq h_f(-x)$ for all $x\in \calX$. Therefore, for any $ f,f'\in \calF$, $h_f(x) \neq h_{f'}(x)$ if and only if $h_f(-x) \neq h_{f'}(-x)$, directly implying our claim. 

    Now, if we consider a subset $Q\subseteq \calX$ of the form given above, we have that the optimal regression function on this subset is given by:
    \begin{align*}
        w^*_Q[i] = \begin{cases}
            \left(\tfrac{|Q|}{2}\right)^{-\frac{1}{2}} & \text{if $e_i\in Q$}\\
            0 & \text{otherwise.}
        \end{cases}
    \end{align*}
    and therefore, $(w^*_Q)^Tx = (|Q|/2)^{-1/2}(\vec{1}\cdot x)$ for any $x\in Q$. As such, we have that:
    \begin{align*}
        |f^*_{Q}(x)-\tfrac{1}{2}| = \tfrac{1}{2}\left(\tfrac{|Q|}{2}\right)^{-\tfrac{1}{2}} \geq \tfrac{1}{2}d^{-1/2} = \psi(|\eta(x)-\tfrac{1}{2}|),
    \end{align*}
    where the inequality holds true because $|Q|\leq 2d$ by construction and the final equality holds true because $|\eta(x)-\tfrac{1}{2}|=\tfrac{1}{2}$ by construction as well. 

    Alternatively, since $\eta(x) = 1$ exactly when $f^*(x) = \tfrac{1}{2}+\tfrac{1}{2}d^{-1/2}$ and $\eta(x) = 0$ exactly when $f^*(x) = \tfrac{1}{2}-\tfrac{1}{2}d^{-1/2}$, we have that $|\eta(x)-f^*(x)| = \tfrac{1}{2}-\tfrac{1}{2}d^{-1/2}$ for all $x$ implying our claim on its approximate realizability.
\end{proof}

\section{The Importance of Convexity of the Benchmark Class}\label{convexity discussion}

In this section, we provide a proof that for any strongly convex surrogate loss, convexity of $\calF$ ensures the excess surrogate risk of a function is bounded by its $L_2(\calD_\calX)$ distance to the surrogate risk minimizer in $\calF$. Although the strong convexity constant of the squared loss is 1, for this case the factor of two can be shaved off (see Lemma 1 of \cite{foster2023foundationsreinforcementlearninginteractive} for example).

\begin{lemma}
\label{lem:strongly-convex-excess}
For a convex class $\calF$ and surrogate loss $\ell_{\Phi} : \mathbb{R}^K \times \calY \to \mathbb{R}$ that is $\beta_\Phi$-strongly convex in its first argument, we have, for all $f \in \calF$:
$$
\|f - f^*\|_{\calD_\calX}^2 \leq \frac{2}{\beta_\Phi}\mathcal{E}_{\ell_\Phi}(f,\calF).
$$
\end{lemma}

\begin{proof}
By $\beta_\Phi$-strong convexity of $\ell_\Phi(\cdot, y)$ over all realizable inputs, we have:
$$
\ell_\Phi(f(x), y) \geq \ell_\Phi(f^*(x), y) + \langle \nabla \ell_\Phi(f^*(x), y), f(x) - f^*(x) \rangle + \frac{\beta_\Phi}{2} \|f(x) - f^*(x)\|_2^2.
$$
Taking an expectation over $(x,y) \sim \calD$ gives us:
$$
 \mathcal{E}_{\ell_\Phi}(f,\calF) \geq \mathbb{E}_{(x,y)\sim \calD}\left[ \langle \nabla \ell_\Phi(f^*(x), y), f(x) - f^*(x) \rangle \right] + \frac{\beta_\Phi}{2} \|f - f^*\|_{\calD_\calX}^2.
$$
Since $f^*$ minimizes surrogate risk over the convex set $\calF$, the expectation is nonnegative implying our desired result.
\end{proof}

%% file: main.bbl
\newcommand{\etalchar}[1]{$^{#1}$}
\begin{thebibliography}{FRSLX20}

\bibitem[Aga13]{pmlr-v28-agarwal13}
Alekh Agarwal.
\newblock Selective sampling algorithms for cost-sensitive multiclass prediction.
\newblock In Sanjoy Dasgupta and David McAllester, editors, {\em Proceedings of the 30th International Conference on Machine Learning}, volume~28 of {\em Proceedings of Machine Learning Research}, pages 1220--1228, Atlanta, Georgia, USA, 17--19 Jun 2013. PMLR.

\bibitem[AT07]{audibert_2007}
Jean-Yves Audibert and Alexandre~B. Tsybakov.
\newblock Fast learning rates for plug-in classifiers.
\newblock {\em The Annals of Statistics}, 35(2), April 2007.

\bibitem[BBL06]{balcan2006agnostic}
Maria-Florina Balcan, Alina Beygelzimer, and John Langford.
\newblock Agnostic active learning.
\newblock In {\em Proceedings of the 23rd International Conference on Machine Learning}, pages 65--72, 2006.

\bibitem[BJM06]{bartlett2006convexity}
Peter~L. Bartlett, Michael~I. Jordan, and Jon~D. McAuliffe.
\newblock Convexity, classification, and risk bounds.
\newblock {\em Journal of the American Statistical Association}, 101(473):138--156, 2006.

\bibitem[Bou02]{1444}
O.~Bousquet.
\newblock {\em Concentration Inequalities and Empirical Processes Theory Applied to the Analysis of Learning Algorithms}.
\newblock PhD thesis, Biologische Kybernetik, 2002.

\bibitem[CN08]{castro2008minimax}
Rui~M. Castro and Robert~D. Nowak.
\newblock Minimax bounds for active learning.
\newblock {\em IEEE Transactions on Information Theory}, 54(5):2339--2353, 2008.

\bibitem[Das04]{dasgupta_2004}
Sanjoy Dasgupta.
\newblock Analysis of a greedy active learning strategy.
\newblock In L.~Saul, Y.~Weiss, and L.~Bottou, editors, {\em Advances in Neural Information Processing Systems}, volume~17. MIT Press, 2004.

\bibitem[Das05]{dasgupta_2005}
Sanjoy Dasgupta.
\newblock Coarse sample complexity bounds for active learning.
\newblock In Y.~Weiss, B.~Sch\"{o}lkopf, and J.~Platt, editors, {\em Advances in Neural Information Processing Systems}, volume~18. MIT Press, 2005.

\bibitem[FR20]{foster2020beyond}
Dylan~J. Foster and Alexander Rakhlin.
\newblock Beyond ucb: Optimal and efficient contextual bandits with regression oracles.
\newblock In {\em Proceedings of the 37th International Conference on Machine Learning}, pages 3199--3210, 2020.

\bibitem[FR23]{foster2023foundationsreinforcementlearninginteractive}
Dylan~J. Foster and Alexander Rakhlin.
\newblock Foundations of reinforcement learning and interactive decision making, 2023.

\bibitem[FRSLX20]{Foster2020}
Dylan~J. Foster, Alexander Rakhlin, David Simchi-Levi, and Yunzong Xu.
\newblock Instance-dependent complexity of contextual bandits and reinforcement learning: A disagreement-based perspective.
\newblock {\em ArXiv}, abs/2010.03104:6, 2020.

\bibitem[FSST97]{freund_1997}
Yoav Freund, {H. Sebastian} Seung, Eli Shamir, and Naftali Tishby.
\newblock Selective sampling using the query by committee algorithm.
\newblock {\em Machine Learning}, 28(2-3):133--168, 1997.

\bibitem[Han07]{hanneke2007bound}
Steve Hanneke.
\newblock A bound on the label complexity of agnostic active learning.
\newblock In {\em Proceedings of the 24th International Conference on Machine Learning ({ICML} 2007)}, pages 353--360, 2007.

\bibitem[Han09]{hanneke2009adaptive}
Steve Hanneke.
\newblock Adaptive rates of convergence in active learning.
\newblock In {\em Proceedings of the 22nd Annual Conference on Learning Theory ({COLT} 2009)}, pages 249--264, 2009.

\bibitem[Han11]{hanneke2011rates}
Steve Hanneke.
\newblock Rates of convergence in active learning.
\newblock {\em The Annals of Statistics}, 39(1):333--361, 2011.

\bibitem[Han14]{hanneke2014theory}
Steve Hanneke.
\newblock Theory of disagreement-based active learning.
\newblock {\em Foundations and Trends® in Machine Learning}, 7(2-3):131--309, 2014.

\bibitem[Han17]{hanneke2017nonparametric}
Steve Hanneke.
\newblock Nonparametric active learning, part 1: Smooth regression functions, 2017.

\bibitem[Hau92]{HAUSSLER89}
David Haussler.
\newblock Decision theoretic generalizations of the pac model for neural net and other learning applications.
\newblock {\em Information and Computation}, 100(1):78--150, 1992.

\bibitem[Hau95]{HAUSSLER95}
David Haussler.
\newblock Sphere packing numbers for subsets of the boolean n-cube with bounded vapnik-chervonenkis dimension.
\newblock {\em Journal of Combinatorial Theory, Series A}, 69(2):217--232, 1995.

\bibitem[Hsu10]{hsu2010algorithms}
Daniel~J. Hsu.
\newblock {\em Algorithms for Active Learning}.
\newblock PhD thesis, University of California, San Diego, 2010.

\bibitem[HY15]{hanneke_2015}
Steve Hanneke and Liu Yang.
\newblock Minimax analysis of active learning.
\newblock {\em Journal of Machine Learning Research}, 16(109):3487--3602, 2015.

\bibitem[HY19]{hanneke_2019}
Steve Hanneke and Liu Yang.
\newblock Surrogate losses in passive and active learning.
\newblock {\em Electronic Journal of Statistics}, 13(2), January 2019.

\bibitem[KAH{\etalchar{+}}21]{krishnamurthy}
Akshay Krishnamurthy, Alekh Agarwal, Tzu-Kuo Huang, Hal~Daume III, and John Langford.
\newblock Active learning for cost-sensitive classification, 2021.

\bibitem[Kol10]{koltchinskii2009rademacher}
Vladimir Koltchinskii.
\newblock Rademacher complexities and bounding the excess risk in active learning.
\newblock {\em Journal of Machine Learning Research}, 11:2457--2485, 2010.

\bibitem[KYZ22]{kpotufe2022nuancesmarginconditionsdetermine}
Samory Kpotufe, Gan Yuan, and Yunfan Zhao.
\newblock Nuances in margin conditions determine gains in active learning, 2022.

\bibitem[Kä06]{kaariainen2006active}
Matti Kääriäinen.
\newblock Active learning in the non-realizable case.
\newblock In {\em Algorithmic Learning Theory, 17th International Conference, {ALT} 2006}, volume 4264 of {\em Lecture Notes in Computer Science}, pages 63--77. Springer, 2006.

\bibitem[LBW98]{lee1998importance}
Wee~Sun Lee, Peter~L. Bartlett, and Robert~C. Williamson.
\newblock The importance of convexity in learning with squared loss.
\newblock {\em IEEE Transactions on Information Theory}, 44(5):1974--1980, 1998.

\bibitem[LCK17]{locatelli2017adaptivitynoiseparametersnonparametric}
Andrea Locatelli, Alexandra Carpentier, and Samory Kpotufe.
\newblock Adaptivity to noise parameters in nonparametric active learning, 2017.

\bibitem[LRS15]{liang2015learning}
Tengyuan Liang, Alexander Rakhlin, and Karthik Sridharan.
\newblock Learning with square loss: Localization through offset rademacher complexity.
\newblock In {\em Proceedings of The 28th Conference on Learning Theory}, pages 1260--1285, 2015.

\bibitem[Min12]{minsker2012plugin}
Stanislav Minsker.
\newblock Plug-in approach to active learning.
\newblock {\em Journal of Machine Learning Research}, 13:67--90, 2012.

\bibitem[MN06]{Massart_2006}
Pascal Massart and {\'E}lodie Nédélec.
\newblock Risk bounds for statistical learning.
\newblock {\em The Annals of Statistics}, 34(5), October 2006.

\bibitem[Pol84]{pollard1984convergence}
David Pollard.
\newblock {\em Convergence of Stochastic Processes}.
\newblock Springer-Verlag, New York, 1984.

\bibitem[RST17]{Rakhlin_2017}
Alexander Rakhlin, Karthik Sridharan, and Alexandre~B. Tsybakov.
\newblock Empirical entropy, minimax regret and minimax risk.
\newblock {\em Bernoulli}, 23(2), May 2017.

\bibitem[SSSW23]{sekhari2023selective}
Ayush Sekhari, Karthik Sridharan, Wen Sun, and Runzhe Wu.
\newblock Selective sampling and imitation learning via online regression.
\newblock In {\em Advances in Neural Information Processing Systems}, 2023.

\bibitem[Tsy04]{tsybakov2004optimal}
Alexander~B. Tsybakov.
\newblock Optimal aggregation of classifiers in statistical learning.
\newblock {\em The Annals of Statistics}, 32(1):135--166, 2004.

\bibitem[ZN22]{zhu2022efficientactivelearningabstention}
Yinglun Zhu and Robert Nowak.
\newblock Efficient active learning with abstention, 2022.

\end{thebibliography}
